\documentclass[twoside,11pt]{article}

\usepackage{lmodern}
\usepackage{graphicx}
\usepackage{multirow,tabularx}
\usepackage{float}
\usepackage[point]{fltpoint}
\usepackage{todonotes}
\usepackage{textcomp}
\usepackage{amsmath,amssymb}
\usepackage{ifthen}
\usepackage{color}
\usepackage{xcolor}
\usepackage{tabularx}
\usepackage{booktabs}
\usepackage{longtable}
\usepackage{footmisc}
\usepackage{rotating}

\usepackage[listofformat=parens, justification=centering, subrefformat=subparens]{subfig}  % have multiple floats in one figure

\DeclareGraphicsExtensions{.pdf,.jpg,.png}
\graphicspath{{./graphics/}{./figures/}}

\newcolumntype{C}{X<{\centering}}

\definecolor{lightgreen}{rgb}{0.67, 0.88, 0.69}
\definecolor{darkgreen}{rgb}{0,0.55,0}
\definecolor{linkcolor}{rgb}{0,0,.65}

\newcommand\ig[2][1]{\includegraphics[width=#1\textwidth, page=1]{#2}}

\newcommand\Caption[3][]{\caption[#2]{\label{#1}\textsc{#2}. \small#3}}

\newcommand\ie{\textit{i.\,e.}}
\newcommand\eg{\textit{e.\,g.}}
\newcommand\proser{PROSER$^*$}

\renewcommand\sec[1]{Section~\ref{sec:#1}}
\newcommand\fig[1]{Figure~\ref{fig:#1}}
\newcommand\sfig[1]{Figure~\subref{fig:#1}}
\newcommand\tab[1]{Table~\ref{tab:#1}}

\newcommand\T{\ensuremath{^\mathrm{T}}}

\DeclareMathOperator*{\argmin}{arg\,min}
\DeclareMathOperator*{\argmax}{arg\,max}

\newcommand{\p}[1]{\ensuremath{P_{#1}}}
\newcommand{\td}[2][Manuel]{\todo[inline]{\textit{#1:} #2}}

\newcommand\new[2][M]{#2}

\usepackage[preprint]{jmlr2e}

\usepackage{lastpage}
\jmlrheading{}{}{}{}{}{}{Bisgin, Palechor, Suter, and G\"unther}

\begin{document}

\title{Large-Scale Evaluation of \\ Open-Set Image Classification Techniques}

\author{\name Halil Bisgin \email bisgin@umich.edu \\
       \addr Department of Computer Science\\
       University of Michigan-Flint\\
       Flint, MI 48502, USA\\
       \AND
       \name Andres Palechor \email andrespalechor11@gmail.com \\
       \name Mike Suter \email mike.milow.suter@gmail.com \\
       \name Manuel G\"unther \email siebenkopf@googlemail.com \\
       \addr Department of Informatics\\
       University of Zurich\\
       Andreasstrasse 15\\
       8050 Zurich, Switzerland}

\editor{TBA}

\maketitle

\begin{abstract}%   <- trailing '%' for backward compatibility of .sty file
The goal for classification is to correctly assign labels to unseen samples.
However, most methods misclassify samples with unseen labels and assign them to one of the known classes.
Open-Set Classification (OSC) algorithms aim to maximize both closed and open-set recognition capabilities.
Recent studies showed the utility of such algorithms on small-scale data sets, but limited experimentation makes it difficult to assess their performances in real-world problems.
Here, we provide a comprehensive comparison of various OSC algorithms, including training-based (SoftMax, Garbage, EOS) and post-processing methods (Maximum SoftMax Scores, Maximum Logit Scores, OpenMax, EVM, PROSER), the latter are applied on features from the former.
We perform our evaluation on three large-scale protocols that mimic real-world challenges, where we train on known and negative open-set samples, and test on known and unknown instances.
Our results show that EOS helps to improve performance of almost all post-processing algorithms.
Particularly, OpenMax and PROSER are able to exploit better-trained networks, demonstrating the utility of hybrid models.
However, while most algorithms work well on negative test samples -- samples of open-set classes seen during training -- they tend to perform poorly when tested on samples of previously unseen unknown classes, especially in challenging conditions.

\end{abstract}


\section{Introduction}
\label{sec:introduction}

The automated classification of visual objects into several classes has a long history in science.
Specifically, categorical classification is trying to discriminate between different categories, such as identifying handwritten digits, traffic signs, birds, or broader categories as provided by the ImageNet data set \citep{deng2009imagenet}.
The breakthrough in this image classification task was by introducing convolutional neural networks to perform this task \citep{lecun1995learning,krizhevsky2012alexnet}, which has shown to outperform any traditional method.
Nowadays, more modern and deeper network architectures are created to perform this task \citep{he2016deep,huang2017densenet,yuan2021tokens}, which are typically trained on large amounts of data with (variations of) the SoftMax loss.

When during deployment only samples of the learned classes are present, these classifiers perform \emph{closed-set} classification.
However, when applying these classifiers in the real world, it cannot be guaranteed that the input always comes from previously seen classes.
For example, when a digit classifier is presented with a letter, it will assign one of the ten digits instead \citep{dhamija2018agnostophobia}.
There are several approaches to counter this behavior.
One such class of methods is \emph{anomaly} or \emph{out-of-distribution detection}, which will classify whether a certain object is likely to belong to a class unseen during training.
Such methods usually perform well when the unknown test data is far away from the known classes, but will typically fail when known and unknown classes are similar.
One easy \emph{post-processing} method for anomaly detection is the Maximum SoftMax Score (MSS) approach by \cite{hendrycks2017baseline}, which turns the network outputs into probabilities through SoftMax activation and thresholds the maximum achieved probability.
An alternative is Maximum Logit Score (MLS) approach from \cite{hendrycks2022scaling}, which directly employs the network outputs (also known as the \emph{logits}) as confidence scores without relying on SoftMax transformation.

Another class of methods deals with the issue of unknown test samples via Open-Set Classification (OSC).
Here, a classifier that is able to classify the known classes is augmented with the possibility to reject unknown samples.
While early OSC approaches, such as OpenMax \citep{bendale2016openmax} and Extreme Value Machine (EVM) \citep{rudd2017evm} mainly rely on output of trained networks in the context of deep learning, later methods try to modify the network such that it has an option to reject unknown samples.
One typical example for the latter group is introducing an additional \emph{garbage class} as an output to the network, an approach that arose early \citep{matan1990handwritten}, was applied in many object detection methods \citep{ren2015faster,redmon2016yolo,jiang2017face,zhao2019object}, and now has become widely adopted in open-set classification \citep{ge2017generative,neal2018counterfactual,zhou2021placeholders,chen2022reciprocal}.
A different approach was introduced by \cite{dhamija2018agnostophobia} who proposed the Entropic Open-Set (EOS) loss and the Objectosphere loss to avoid the network to produce any large probability for unknown samples, without needing to model the probability of unknown explicitly.

%. Typically, an additional \emph{garbage class} is added as an output to the network,

%Two kinds of approaches exist in the literature: (a) \emph{feature-based methods} use existing features, which are nowadays typically extracted from a pre-trained closed-set network, and learn a secondary classifier to include a reject option, and (b) \emph{training-based methods} try improving the network by including a reject option into the training procedure of the network directly.

%Most of the early approaches to open-set classification fall into the first category.
%\cite{bendale2016openmax} introduced the OpenMax method to add a probability of unknown by modeling deep feature distributions.
%\cite{rudd2017evm} introduced the Extreme Value Machine (EVM) to compute a probability of sample inclusion for each of the known classes based on their %feature similarities and showed its extensibility to deep feature space.
%\cite{guenther2020watchlist} implemented a shallow network on top of deep features extracted from a face recognition network to perform open-set face %recognition.
%A clear disadvantages of these approaches is that they cannot undo any damage to the feature representation learned by the pre-trained network, e.g., when %samples from known and unknown classes overlap in deep feature space, neither of them can improve the open-set performance of the network.

%Many of the more modern techniques try to modify the network such that it has an option to reject unknown samples.

All these types of network-based OSC algorithms require training on \emph{negative} data, also termed \emph{known unknown} data in the literature, \ie~data that does not belong to any of the known classes.
Some approaches made use of real negative samples from different data sets \citep{dhamija2018agnostophobia,palechor2023protocols}, while other researchers tried to build such negative samples by modifying or combining known samples \citep{zhou2021placeholders,wilson2023safe} or by artificially generating negative samples \citep{ge2017generative}.
For example, the PROSER algorithm \citep{zhou2021placeholders} performs manifold mixup of middle-level features of two samples form different known classes \citep{verma2019manifold} to build negative samples.

Many evaluation data sets in the open-set classification task are small, with small numbers of classes, but an abundance of samples per class.
For example, many researchers in open-set image classification make use of (various combinations of) MNIST \citep{lecun1998mnist}, CIFAR-10/100 \citep{krizhevsky2009cifar}, SVHN \citep{netzer2011svhn}, Fashion-MNIST \citep{xiao2017fashion} or Tiny ImageNet challenge \citep{le2015tiny}, but other small-scale data sets are also used.
Some researchers even include random noise with different distributions as unknown samples \citep{liang2017odin}, which clearly does not provide any insight on how algorithms react to samples of unknown classes.
Generally, these data sets have been introduced for closed-set image classification, and different random and non-standard evaluation protocols are designed by many researchers.
While designing new open-set techniques might be fostered by such small-scale data sets since experimentation is quick, they do not represent realistic views of real-world tasks.
The issue with these protocols is that known and unknown classes are typically visually and semantically distinctive and, thus, the task at hand is more related to out-of-distribution detection than on open-set classification.
The lack of large-scale evaluations has also been acknowledged in the out-of-distribution detection field, for which lately new evaluation benchmarks have been introduced by \cite{yang2022openood}, but their proposed OSC benchmark still relies on small-scale data sets.

Additionally, most researchers evaluate their methods with metrics that are not designed to (and, therefore, not able to) evaluate open-set classification realistically.\footnote{See \cite{dhamija2018agnostophobia,boult2019learning,wang2022openauc,palechor2023protocols} for a more detailed discussion on drawbacks of existing evaluation techniques.}
\cite{dhamija2018agnostophobia,dhamija2019improving} have designed the Open-Set Classification Rate (OSCR) curve, which treats known and unknown test samples differently and follows the well-known Open-Set ROC curve actively applied in face recognition \citep{phillips2011evaluation,nist2022ongoing}.
The OSCR metric is nowadays widely adopted for open-set classification, but in many publications it is used as a single number \citep{vaze2022openset,chen2022reciprocal} without specifying how exactly this number is computed, so a comparison with these works is not possible -- we here assume that they report the Area Under the OSCR curve.

\begin{figure}[tb]
  \ig{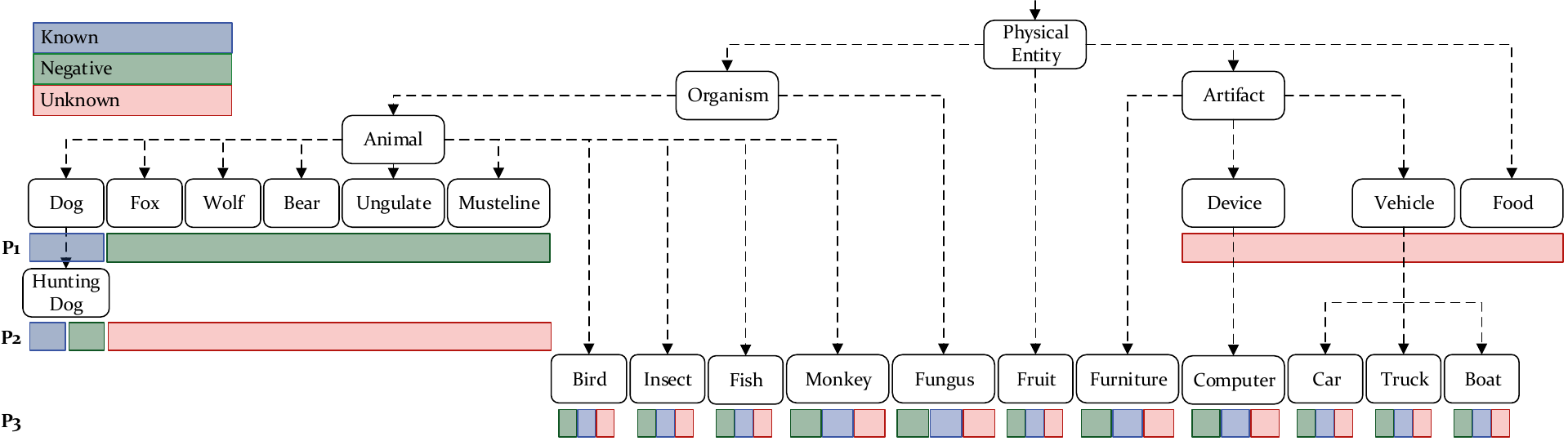}
  \Caption[fig:protocols]{Open-Set Protocols for ImageNet}{
      This figure shows the partition of classes into known, negative and unknown within the three different protocols, \p1, \p2, and \p3.
      By following the WordNet hierarchy \citep{miller1998wordnet} which is shown with the dashed lines indicating an ``is-a'' relationship, we sample our final classes from the leaf nodes of the intermediate-level superclasses named above the colored bars.
      The colored bars below indicate that its subclasses are sampled for the same color codes representing knowns, negatives and unknowns.
      For example, all subclasses of "Dog" are used as known classes in \p1, while the subclasses of "Hunting Dog" are partitioned into knowns and negatives in \p2.
      On the other hand, \p3 has several intermediate nodes that are partitioned into known, negative, and unknown classes.
      Those partitions also constitute training, validation, and test sets in each protocol.
      While known and negative classes are available during training and validation, unknown classes only appear in test time.
      More details on the protocols are provided by \cite{palechor2023protocols}.
    }
\end{figure}

Lately, we have introduced a new open-set image classification challenge \citep{palechor2023protocols} based on the ImageNet data set used in the ImageNet Large Scale Visual Recognition Challenges (ILSVRC) \citep{russakovsky2015imagenet}.
Using the WordNet hierarchy \citep{miller1998wordnet}, we designed three different evaluation protocols with varying semantic similarity between known and unknown classes, as can be seen in \fig{protocols}.
Additionally, we provide real negative samples to foster research in training OSC methods with them.
The variations in semantic similarities, which can be also inferred by following the dashed lines in the same figure, contributed to the difficulty levels of protocols.
For instance, in the first protocol, \p1, known and unknown classes are semantically quite distant from each other while negative classes are close to the known class, which made \p1 easy for open-set and hard for closed-set classification.
Protocols \p2 and \p3 are also constructed by following different semantic distances to have richer and more realistic scenarios.
While other evaluation protocols on ImageNet exist \citep{bendale2016openmax,vaze2022openset} they make use of the total of the ILSVRC 2012 data set as known classes and do not provide negative samples, which disallows to have large shifts between known and unknown classes, limits the algorithms that can be applied to them, and prohibit extensive experimentation through the expensive training on the large data set.

The contributions of this paper are as follows:

\begin{itemize}
  \item We perform the first large-scale evaluation of several training-based and post-processing methods for open-set classification.
  \item For the first time in the literature, we combine these orthogonal methods to further improve performance.
  \item We exploit the evaluation protocols that we previously defined \citep{palechor2023protocols} and the evaluation metric designed by \cite{dhamija2018agnostophobia} to perform a fair and realistic evaluation of the methods.
  \item We make use of publicly available implementations as far as possible, and re-implement methods when required. We provide all of our implementations and evaluation techniques as open-source package,\footnote{\label{fn:package}\url{https://github.com/AIML-IfI/openset-imagenet-comparison}} hoping to foster more research and more fair and realistic evaluations of open-set classification techniques in the future.
\end{itemize}

\section{Open-Set Classification Taxonomy}
\label{sec:taxonomy}
As we noted in Section \ref{sec:introduction}, open-set classification approaches can either perform additional training with some adjustments or make use of deep features to correctly classify known samples and identify unknowns in a deep learning framework.
These two kinds of approaches lead us to a taxonomy we adopt in this work: (a) \emph{training-based methods} try improving the network by including a reject option into the training procedure of the network directly, and (b) \emph{post-processing methods} use existing features or scores, which are nowadays typically extracted from a pretrained closed-set network, to provide probabilities $p_c$ for all \emph{known} classes $c\in\{1,\ldots,K\}$, possibly by learning a secondary classifier.

These probabilities can finally be evaluated in open-set metrics, which we define in \sec{metrics}.
Please note that we do not require to compute the probability of the unknown class $p_{K+1}$.
A low probability for the unknown class $p_{K+1}$ does not indicate a high probability for any of the known classes.
On the other hand, due to SoftMax requiring that probabilities sum up to 1, a large $p_{K+1}$ probability enforces low probabilities for all known classes.
Therefore, $p_{K+1}$ does not add any new information.

\subsection{Training-based Methods}
A typical processing with deep networks takes place as shown in \fig{processing}, where images are passed to a backbone network, which is ResNet-50 in our case, and their logits $\vec z$ output from a fully-connected layer are turned into probabilities $\vec y$ through SoftMax:
\begin{equation}
  \label{eq:softmax}
  y_{n,c} = \frac{e^{z_{n,c}}}{\sum\limits_{c'=1}^C e^{z_{n,c'}}}\,.
\end{equation}
Based on these, the weighted categorical cross-entropy loss can be computed as:
\begin{equation}
  \label{eq:cce}
  \mathcal J_{\mathrm{CCE}} = \new{-}\frac1N \sum\limits_{n=1}^N \sum\limits_{c=1}^C w_c t_{n,c} \log y_{n,c}
\end{equation}
where $N$ is the number of samples in our data set (note that we utilize batch processing), $t_{n,c}$ is the target label of the $n$th sample for class $c$, $w_c$ is a class-weight for class $c$ and $y_{n,c}$ is the confidence of class $c$ for sample $n$ using SoftMax activation.

\begin{figure}[t]
  \small\centering
  \begin{tikzpicture}[scale=.5]
    \node[draw](Image) at (0,0) {\includegraphics[width=.1\textwidth]{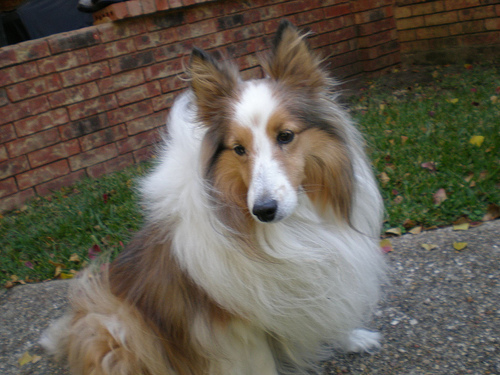}};
    \node[draw,rectangle](Magic) at (10,0) {
      \begin{minipage}{.4\textwidth}\Huge \centering \textsc{Magic}$^*$\\[3ex] \small $^*$following Arthur C. Clarke's 3rd law:\\[1ex] ``Any Sufficiently Advanced Technology is Indistinguishable from Magic.''\end{minipage}
%      \begin{minipage}{.4\textwidth}\huge \centering Backbone Network\\[1ex] \normalsize Convolutions\\Pooling\\\ldots\\(ResNet-50 in our case)\end{minipage}
    };
    \node[draw,rectangle](Features) at (19,0) {\begin{turn}{270} Deep Features $\vec \varphi$\end{turn}};
    \node[draw,rectangle](Logits) at (22,0) {\begin{turn}{270} Logits $\vec z$\end{turn}};
    \node[draw,rectangle](Probabilities) at (25,0) {\begin{turn}{270} Probabilities $\vec y$\end{turn}};

    \draw[->,thick] (Image) -> (Magic);
    \draw[->,thick] (Magic) -> (Features);
    \draw[->,thick] (Features) -> node [above] {\begin{turn}{270} Linear \end{turn}} node [below]{\begin{turn}{270} $\mathbf W$\end{turn}} (Logits);
    \draw[->,thick,dashed] (Logits) -> node [above] {\begin{turn}{270} SoftMax\end{turn}} (Probabilities);
  \end{tikzpicture}
  \Caption[fig:processing]{Processing with Deep Networks}{An image is presented to the backbone network, which extracts deep features $\vec\varphi$ that are then processed with a Linear layer to logits $\vec z$, and further with SoftMax to probabilities $\vec y$.}
\end{figure}

In our training-based category, we use three variations of this paradigm, which employ SoftMax, Garbage class, and Entropic Open-Set (EOS) loss.
These are the same three techniques to train the deep networks as in \citep{palechor2023protocols}.
We solely use categorical cross-entropy loss on top of SoftMax activations (also known as the SoftMax loss) to train our networks.
The three different training approaches differ with respect to the targets $t_{n,c}$ and the weights $w_c$, and how negative samples are handled.
A more technical explanation of these methods are given in \sec{training-based}.

While the largest amount of research in open-set classification tries to artificially generate effective negative samples to train the networks \citep{ge2017generative,neal2018counterfactual,chen2020learning,chen2022reciprocal}, in our evaluation we solely rely on the real negative samples provided in our protocols.
For a more detailed analysis of other existing OSC techniques, we refer the reader to \cite{geng2021recent}.

\subsection{Post-processing Methods}
Most of the early approaches to open-set classification fall into the category of feature or score post-processing.
\cite{bendale2016openmax} introduced the OpenMax method to add a probability of unknown by modeling deep feature distributions.
\cite{rudd2017evm} introduced the Extreme Value Machine (EVM) to compute a probability of sample inclusion for each of the known classes based on their feature similarities and \cite{guenther2017toward} showed its extensibility to deep feature space.
\cite{guenther2020watchlist} implemented a shallow network on top of deep features extracted from a face recognition network to perform open-set face recognition.
Recently, Hendrycks and colleagues \citep{hendrycks2017baseline,hendrycks2022scaling} indicated that simply thresholding the output of the pretrained network can provide a reasonable option for OSC.

A clear disadvantages of these approaches is that they cannot undo any damage to the feature representation learned by the pretrained network, \eg, when samples from known and unknown classes overlap in deep feature space, neither of them can improve the open-set performance of the network.
Therefore, \cite{zhou2021placeholders} have introduced a hybrid approach, the PlaceholdeRs for Open-SEt Recognition (PROSER), which starts with a pretrained network and adds post-hoc open-set capabilities to that network.

Due to the ease of application or the availability of open source code, our post-processing category includes the following five open-set methods on top of our trained networks for which we provide formal definitions in \sec{post-processing}:

\begin{itemize}
\item Maximum SoftMax Score (MSS) uses the SoftMax scores \eqref{eq:softmax} of the networks and serves as the baseline \citep{hendrycks2017baseline}.
\item Maximum Logit Score (MLS) relies on the logits $z_c$, \ie, the direct outputs of the network \citep{hendrycks2022scaling}.
\item OpenMax uses logits and deep features to obtain a probability of unknown \citep{bendale2016openmax}.
\item Extreme Value Machine (EVM) exploits deep features to reject unknown samples \citep{rudd2017evm}.
\item PlaceholdeRs for Open-SEt Recognition (PROSER) fine-tunes the network and adds outputs for unknown \citep{zhou2021placeholders}.
\end{itemize}

%

%Many of the more modern techniques try to modify the network such that it has an option to reject unknown samples.

\section{Evaluated Methods}
\label{sec:methods_new}
We evaluate two complementary types of open-set methods: training-based and post-processing.
Additionally, for the first time in the literature, we try combinations of training-based and post-processing methods.
We first train the training-based method to provide features that are able to separate known classes from each other and also from unknown classes, and we apply post-processing algorithms to further improve the separability.
Here, we provide the technical details of all applied algorithms.

\subsection{Training-based Methods}
\label{sec:training-based}
The first approach in the training-based category is the plain SoftMax loss (S) that is trained only on samples from the $K$ known classes, since this has been shown to be a strong baseline for out-of-distribution detection algorithms \citep{hendrycks2017baseline,hendrycks2022scaling}.
The number of network outputs $C=K$ is equal to the number of known classes, and the targets are computed as one-hot encoding:
\begin{equation}
  \label{eq:targets-one-hot}
  \forall n, c\in\{1,\ldots,C\}: t_{n,c} = \begin{cases}1 & c = \tau_n \\ 0 & \text{otherwise} \end{cases}
\end{equation}
where $1\leq\tau_n\leq K$ is the class label of the sample $n$.
Following standard procedure for training deep networks on ILSVRC that contains almost balanced training classes, we select the weights for each class to be identical: $\forall c: w_c = 1$.

The second approach is often found in object detection models \citep{dhamija2020elephant} which collect a lot of negative samples from the background of the training images.
Similarly, this approach is used in other methods for open-set learning, such as G-OpenMax \citep{ge2017generative}.\footnote{While these methods try to sample better negatives for training, they rely on this additional class for unknown samples.}
In this Garbage class approach, negative data is used to train an additional network output $z_{K+1}$, so that we have a total of $C=K+1$ outputs.
Since the number of negative samples is usually higher than for known classes, we use class weights to balance them:
\begin{equation}
  \label{eq:weights-bg}
  \forall c\in\{1,...,C\}: w_c = \frac{N}{CN_c}
\end{equation}
where $N_c$ is the number of training samples for class $c$.
Finally, we use one-hot encoded targets $t_{n,c}$ according to \eqref{eq:targets-one-hot}, including label $\tau_n=K+1$ for negative samples.

Finally, we employ the Entropic Open-Set (EOS) loss \citep{dhamija2018agnostophobia}, which is a simple extension of the SoftMax loss.
Similar to our first approach, we have one output for each of the known classes: $C=K$.
For known samples, we employ one-hot-encoded target values according to \eqref{eq:targets-one-hot}, whereas for negative samples we use identical target values:
\begin{equation}
  \label{eq:targets-equal}
  \forall n, c\in\{1,\ldots,C\}: t_{n,c} = \frac1C
\end{equation}
Sticking to the implementation of \cite{dhamija2018agnostophobia}, we select the class weights to be $\forall c\colon w_c = 1$ for all classes including the negative class, and leave the optimization of these values for future research.
For the negative samples required by Garbage and EOS, we solely rely on the negative classes defined in the evaluation protocols (cf.~\sec{protocols}), but we acknowledge that different negative samples, such as generative approaches \citep{ge2017generative}, counterfactual images \citep{neal2018counterfactual}, noisy samples \citep{wilson2023safe}, manifold mix-up \citep{verma2019manifold} or other new approaches to obtain negative samples might further improve the training of the networks.

\subsection{Post-processing Methods}
\label{sec:post-processing}

Our first post-processing method is based on \cite{hendrycks2017baseline}, which we call Maximum SoftMax Score (MSS).
It is also our baseline approach that relies on the scores or probabilities derived from \eqref{eq:softmax}.
It basically acts on the output of the trained networks defined above and helps in the decision-making based on the probability for the class: $p_c = y_c$.
The second approach following the idea from \cite{hendrycks2022scaling}, which we coin as Maximum Logit Score (MLS), on the other hand, does not push the network outputs through SoftMax.
Instead, it works with the logits directly and considers the logit values $p_c = z_c$.
Please note that these logit scores are not direct probability values, but none of our evaluation metrics in \sec{metrics} actually requires probabilities.

%\item Maximum SoftMax Score (MSS) which uses the scores of the SoftMax layer \eqref{eq:softmax} of the networks as defined in the previous section and serves as the baseline .
%\item Maximum Logit Score (MLS) which uses the direct outputs of the network or logits $z_c$ \citep{hendrycks2022scaling}.

%The second method follows the idea from \cite{hendrycks2022scaling} where the logits $z_c$, \ie, the direct outputs of the network are used instead of pushing them through SoftMax, and we term this method .

As a third method, we make use of the original OpenMax technique as introduced by \cite{bendale2016openmax}, using a slight twist as implemented the VAST software package.\footnote{\label{fn:vast}\url{https://github.com/Vastlab/vast}}
OpenMax makes use of deep features $\vec\varphi$ extracted from the penultimate layer of the deep network, \ie, the layer before the last fully-connected layer as shown in \fig{processing} with weight matrix $\mathbf W$ turns the deep features into logits:
\begin{equation}
  \label{eq:deep-features}
  \vec z = \mathbf W \vec\varphi
\end{equation}
In the original OpenMax paper, \cite{bendale2016openmax} called the deep features Activation Vectors (AV).
For each known class $1\leq c\leq K$, the Mean Activation Vector (MAV) $\vec\mu_c$ is computed by averaging the deep features extracted from all correctly classified known training samples of class $c$.
Additionally, the cosine distances of the MAV $\vec\mu_c$ to all the AVs $\vec\varphi_{n,c}$ of the same class are computed.
Here, we make use of the twist implemented in the VAST package: instead of using the original distances to model the distribution, we multiply the cosine distances by a certain factor $\kappa$, which allows modeling more compact class representations:
\begin{equation}
  \label{eq:openmax-distances}
  d_{n,c} = \kappa(1-\cos(\vec\varphi_n, \vec\mu_c))
\end{equation}
The $\lambda$ \emph{largest} distances per class are used to model a Weibull distribution $\Psi_c$.
For a given test sample, these Weibull distributions are taken to estimate a logit $z_{K+1}$ for being unknown, as well as modifying the other logits for the top $\alpha$ classes.
For details on this procedure, please refer to the original paper \citep{bendale2016openmax}.
This additional logit is included into the SoftMax calculation \eqref{eq:softmax}, to obtain the probabilities $p_c = y_c$ for all known classes $c\in\{1,\ldots,K\}$ and the additional unknown class $y_{K+1}$.\footnote{\label{fn:no-unknown-prob}Note that in our evaluation, we ignore the probability for the unknown class returned by OpenMax or \proser. It is important to understand that the \emph{logit} $z_{K+1}$ is not ignored, but lowers all known class probabilities $y_1,\ldots,y_K$ through SoftMax, which enforces probabilities including $y_{K+1}$ to sum up to 1.}

The fourth method that we employ is the Extreme Value Machine (EVM) as introduced by \cite{rudd2017evm}.
Similar to OpenMax, EVM uses Weibull-calibrated scoring to compute a Probability of Sample Inclusion (PSI) for each class, without computing a specific probability for a sample to be unknown.
Contrary to OpenMax, which learns the probability models on distances between samples from the same class, EVM estimates distances of samples between classes.
For a given activation vector $\vec\varphi_{n,c}$ of class $c$, it computes the cosine distances to all samples $\vec\varphi_{n',c'}$ from all other classes $c'\neq c$, which is again modified by a distance multiplier $\kappa$ to allow tighter class bounds \citep{guenther2017toward}:
\begin{equation}
  \label{eq:evm-distances}
  d_{n,n',c} = \kappa(1-\cos(\vec\varphi_{n,c}, \vec\varphi_{n',c'}))
\end{equation}
Obviously, this is an expensive operation in $\mathcal O(N^2)$, but the VAST package\footref{fn:vast} provides an optimized implementation on the GPU for this task.
From all of these distances, the $\lambda$ \emph{smallest} values are used to train a probability of sample \emph{exclusion} using Weibull fitting, which is turned into a probability of sample \emph{inclusion} $\Psi_{n,c}$ by taking its inverse probability.
These are estimated for each training sample.
To reduce computational cost, \cite{rudd2017evm} added the possibility of model reduction via a cover threshold $\omega$, but we do not make use of this option to ensure highest possible performance.
Finally, to compute the probability $p_c$ of a test sample $\vec\varphi$ to belong to a certain class $c$, the maximum over all probabilities associated with this class is taken:
\begin{equation}
  \label{eq:evm-probability}
  p_c(\vec\varphi) = \max\limits_{n\colon\tau_n = c} \Psi_{n,c}(\vec\varphi)
\end{equation}
For EVM and OpenMax, we rely on the original implementation provided by the authors.\footref{fn:vast}

The final method that we evaluate is PlaceholdeRs for Open-SEt Recognition (PROSER) introduced by \cite{zhou2021placeholders}.
It makes use of mid-level features $\vec\varphi$ from pretrained networks to fine-tune the network.
Similar to OpenMax and the Garbage class approach, PROSER tries to estimate an additional logit $z_{K+1}$ into SoftMax \eqref{eq:softmax}.
To train this logit, PROSER does not rely on external negative samples, but it defines negative samples (which the authors call \emph{data placeholders}) by merging mid-level features of the pretrained network extracted from two randomly selected features of different classes, based on the concept of manifold mix-up \citep{verma2019manifold}:
\begin{equation}
  \label{eq:proser-mixup}
  \vec\vartheta' = \beta \vec\vartheta_{n,c} + (1-\beta) \vec\vartheta_{n',c'} \qquad \text{with } c\neq c'
\end{equation}
where $0\leq\beta\leq1$ is randomly drawn from a Beta distribution.
Additionally, instead of requiring all deep feature representations of unknown samples to gather in the same location in feature space, $B$ logit values (that the authors call \emph{classifier placeholders}) are computed for each negative/unknown sample, and the maximum of the $B$ logits is taken to represent the garbage class:
\begin{equation}
  \label{eq:proser-logits}
  z_{K+1} = \max\limits_{1\leq b\leq B} \vec w_b\T \vec \varphi
\end{equation}
with $\vec w_b$ being learnable parameters.
As before, these logits are used to compute SoftMax activation $y_c$ via \eqref{eq:softmax}, and we directly make use of them for evaluation $p_c=y_c$.\footref{fn:no-unknown-prob}
Furthermore, for the known samples a specific loss function is applied encouraging one of the $B$ classifier placeholders to be the second-highest logit \citep{zhou2021placeholders}.
Finally, the whole network is fine-tuned using the original training images from the known classes, and the generated mid-level features as negatives.
Our implementation is inspired by the original code from the authors.\footnote{\label{fn:proser}\url{https://github.com/zhoudw-zdw/CVPR21-Proser}}
However, the original implementation is not flexible enough to change the backbone network, so we needed to rewrite it to adapt it to other network topologies.
We also corrected some mistakes made in the original code basis, for example, we now actually check that the two samples for the mix-up are not coming from the same class.
Our own re-implementation can be found in our source code package.\footref{fn:package}
To indicate these differences to the original code, we mark our reimplementation as \proser.

\section{Data Set and Evaluation Protocols}
\label{sec:dataset}

\subsection{Evaluation Protocols}
\label{sec:protocols}
In our evaluation, we use the ILSVRC 2012 data set \citep{russakovsky2015imagenet}, for which we have developed three different open-set evaluation protocols \citep{palechor2023protocols} as indicated in \fig{protocols}.
The three evaluation protocols differ in terms of difficulty in both closed and open-set classification.
Additionally, negative classes are sampled such that they are similar to the known classes, since unrelated negative classes have been shown to not help in open-set classification \citep{dhamija2018agnostophobia}.
The samples in each protocol are split into training, validation and test splits, where known and negative classes are provided for training, validation and test, while unknown classes only appear during test time.
For more details on the exact composition of the three protocols, please refer to \cite{palechor2023protocols}.

Protocol \p1 is designed to be relatively difficult in closed-set classification as the task is to differentiate between all different types of dogs provided in ILSVRC.
On the other hand, it should be relatively easy for open-set classification because unknown test samples in \p1 come from semantically distant categories such as devices, vehicles and food.
Negative samples include classes similar to dogs, which basically are other four-legged animals.

Protocol \p3 is the exact opposite, it is more easy in closed-set classification since the enclosed known classes are more dissimilar to each other.
At the same time, the unknown classes are semantically similar to the known classes, so it is likely that texture features are shared between known and unknown classes.
Negative samples belong to the same group of categories as known and unknown classes.

Protocol \p2 is designed to be in the middle between \p1 and \p3, as well as to be quite small so that hyperparameter optimization experiments can be conducted on this protocol.
Known and negative classes are samples from the dog classes, while unknown classes include all other four-legged animals contained in ILSVRC 2012.

\subsection{Evaluation Metric}
\label{sec:metrics}
To evaluate the different approaches, we make use of the Open-Set Classification Rate (OSCR) curve, which handles known and negative/unknown samples separately.
Based on a certain threshold $\theta$ and assuming $K$ different known classes, the Correct Classification Rate (CCR) and the False Positive Rate (FPR) are computed as \citep{palechor2023protocols}:
\begin{align}
  \label{eq:ccr}
  \mathrm{CCR}(\theta) &= \frac{\bigl|\{n \mid \tau_n \leq K \wedge \argmax\limits_{1\leq c\leq K} p_{n,c} = \tau_n \wedge p_{n,\tau_n} \geq \theta \}\bigr|}{|N_K|}\\
  \label{eq:fpr}
  \mathrm{FPR}(\theta) &= \frac{\bigl|\{n \mid \tau_n > K \wedge \max\limits_{1\leq c\leq K} p_{n,c} \geq \theta \}\bigr|}{|N_U|}
\end{align}
where $n$ iterates all test samples and $N_K$ and $N_U$ are the total numbers of known and negative/unknown test samples, while a target label $\tau_n \leq K$ indicates a known sample and $\tau_n > K$ refers to a negative or unknown test sample.
%Please note that both metrics only evaluate the scores for known classes, purposefully leaving out the probability of the unknown class in the \emph{garbage class} approach, as well as for OpenMax and PROSER evaluations.

By varying the threshold from the smallest to the highest possible score value, a curve can be drawn, plotting the CCR over the FPR.
Since most applications require very low numbers of false positives, this curve is typically drawn with a logarithmic FPR axis.
For some approaches, many probability scores $p_{n,c}$ reach the maximum value of 1 to any reasonable precision, in which case there is no threshold $\theta$ that would allow computing low FPR values, so that the OSCR curve does not extend further to the left.
This fact also disallows us computing the Area Under the OSCR curve (nevertheless \cite{vaze2022openset,chen2022reciprocal} seem to do exactly that) since the extent of the FPR axis differs between algorithms.
Additionally, in real-world applications a specific threshold $\theta$ is required, which is typically selected based on a certain FPR, but computing AUOSCR disables selecting such a threshold.

\subsection{Single-Valued Evaluation Metric}

Some of our open-set algorithms, such as OpenMax, EVM and \proser, rely on an already optimized network, but they require an additional hyperparameter optimization step onwards.
Particularly, different combinations of algorithm-specific hyperparameters are tried to reach the best prediction performance.
However, defining the best can only be achieved through a single value calculation which can make the comparison across algorithms and different hyperparameter settings possible.

For a fair comparison of various hyperparameter combinations, whose OSCR curves often end at different points on the FPR axis, we use a set of predefined FPR values to extract available CCR values that correspond to those points.
Particularly, we computed CCR different values for FPR $\zeta \in \{10^{-3}, 10^{-2}, 10^{-1}, 10^0\}$.
Since it is not always possible to find the exact FPR value, we make use of the infimum value if such value exists, and compute the sum over their corresponding CCR:
\begin{equation}
  \begin{aligned}
  \label{eq:ccr@fpr}
  \mathrm{CCR@FPR} &= \sum\limits_{\zeta} \begin{cases} \mathrm{CCR}(\theta_{FPR=\zeta}) & \text{if } \theta_{FPR=\zeta} \text{ exists} \\ 0 & \text{otherwise} \end{cases} \\[1ex]
  \theta_{FPR=\zeta} &= \argmin_\theta\ \mathrm{FPR}(\theta) \quad\text{s.t. }  \mathrm{FPR}(\theta) - \zeta \geq 0
  \end{aligned}
\end{equation}
In doing so, we only take into account existing FPRs and prevent our CCR@FPR measure from being affected by some artificial CCR values at non-reachable FPRs.

\subsection{Comparison to Other Metrics}
In many related publications, open-set systems are evaluated by computing the closed-set accuracy and the Area under the Receiver Operating Characteristics (AUROC) curve.
The former only looks into the known classes, and is identical to the CCR at FPR=1.
The latter concerns how well known and unknown classes can be distinguished by computing False Positive Rates according to \eqref{eq:fpr}, as well as the True Positive Rate (TPR):
\begin{equation}
  \label{eq:tpr}
  \mathrm{TPR}(\theta) = \frac{\bigl|\{n \mid \tau_n \leq K \wedge \max\limits_{1\leq c\leq K} p_{n,c} \geq \theta \}\bigr|}{|N_K|}
\end{equation}
The ROC is computed by varying $\theta$, and the area under that curve is determined.
While we \textbf{vigorously discourage} the usage of this metric since it does not evaluate open-set performance (AUROC only evaluates out-of-distribution detection), we include it in our evaluation, solely to show that the usage of that metric \textbf{can lead to wrong conclusions}, cf.~\sec{OSCR-AUROC}.
\section{Experiments}
\label{sec:experiments}

Having both training-based and post-processing methods, we set up a comprehensive experimental design where not only training-based methods are evaluated, but also post-processing methods make use of results generated by training-based methods.
Hence, we perform a total of 15 experiments on each of the three evaluation protocols, which include a cross-combination of three training-based methods: SoftMax, Garbage, and EOS, and five post-processing methods: MSS, MLS, OpenMax, EVM and \proser.

\subsection{Network Training}
We start our experiments by first training ResNet-50 models for each training-based method on all three protocols under the same conditions, \ie~we use same set of parameter values for the optimizer (Adam), number of epochs (120), and learning rate ($10^{-3}$).
%We do not apply early stopping

When training a deep network, typically one chooses the network after that training epoch that produces the best performance according to a certain validation metric.
Unfortunately, there exists no validation metric that would work in all circumstances.
%We, however, always take the network at the end of the 120 training epochs (or 20 fine-tuning epochs for \proser, see below).
While in our previous publication \citep{palechor2023protocols} we proposed a validation metric that looks into confidence values produced by the network for known and negative validation set samples, we have also shown that this metric does not perform well for all different loss functions.
For example, training a network with the SoftMax approach, \ie, excluding negative samples, will produce the best validation score after a few epochs, but this does not correspond to the best network that can be obtained \citep{palechor2023protocols}.
Selecting different validation metrics, however, would have decreased the comparability across methods.
Therefore, we have decided against relying on a validation metric and always make use of the network after 120 training epochs.

In our software package,\footref{fn:package} we compute and report several validation metrics, including the above-mentioned metric, which we observed would be well-suited for approaches that train with real negative samples.
On the other hand, we never experienced any large amount of overfitting to the training data on any type of validation metric reported by our software package.
In future work, we will try to investigate more on reasonable validation metrics that can be applied across different training regimes.

\subsection{Hyperparameter Optimization}
\begin{table}[t!]
  \centering\footnotesize
  \subfloat[OpenMax]{
  \begin{tabular}{|c|c|c|c|c|c|c|}
    \hline
    \bf Protocol & \bf Training & $\lambda$ & $\kappa$ & $\alpha$ & $\Sigma$ \\\hline\hline
\multirow{3}{*}{$P_1$} & Softmax & 1000 & 2.0 & 10 & 1.5823 \\
 & Garbage & 1000 & 1.5 & 5 & 1.6087 \\
 & EOS & 500 & 2.0 & 10 & 2.2867 \\
\hline
\multirow{3}{*}{$P_2$} & Softmax & 1000 & 2.3 & 10 & 1.1703 \\
 & Garbage & 750 & 1.7 & 5 & 1.2638 \\
 & EOS & 500 & 2.0 & 3 & 1.6239 \\
\hline
\multirow{3}{*}{$P_3$} & Softmax & 750 & 2.3 & 5 & 1.2201 \\
 & Garbage & 1000 & 2.3 & 2 & 1.5077 \\
 & EOS & 250 & 2.0 & 10 & 1.9241 \\
\hline

  \end{tabular}}

  \subfloat[EVM]{
  \begin{tabular}{|c|c|c|c|c|c|c|}
    \hline
    \bf Protocol & \bf Training & $\lambda$ & $\kappa$ & $\omega$ & $\Sigma$ \\\hline\hline
\multirow{3}{*}{$P_1$} & Softmax & 1000 & 0.4 & 1 & 1.5068 \\
 & Garbage & 75 & 0.7 & 1 & 1.3909 \\
 & EOS & 1000 & 0.2 & 1 & 1.4321 \\
\hline
\multirow{3}{*}{$P_2$} & Softmax & 300 & 0.3 & 1 & 1.1642 \\
 & Garbage & 1000 & 0.2 & 1 & 1.2164 \\
 & EOS & 150 & 0.2 & 1 & 1.1182 \\
\hline
\multirow{3}{*}{$P_3$} & Softmax & 100 & 0.5 & 1 & 1.3113 \\
 & Garbage & 500 & 0.5 & 1 & 1.4061 \\
 & EOS & 1000 & 0.2 & 1 & 1.3228 \\
\hline

  \end{tabular}}\hspace*{1em}
  \subfloat[\proser]{
    \begin{tabular}{|c|c|c|c|c|c|c|}
      \hline
      \bf Protocol & \bf Training & $B$ & $\Sigma$ \\\hline\hline
\multirow{3}{*}{$P_1$} & Softmax & 100 & 1.1342 \\
 & Garbage & 2 & 1.1445 \\
 & EOS & 5 & 2.3405 \\
\hline
\multirow{3}{*}{$P_2$} & Softmax & 1 & 1.2310 \\
 & Garbage & 5 & 1.4862 \\
 & EOS & 10 & 1.6420 \\
\hline
\multirow{3}{*}{$P_3$} & Softmax & 10 & 1.3456 \\
 & Garbage & 10 & 1.7394 \\
 & EOS & 100 & 1.7226 \\
\hline

    \end{tabular}}

  \Caption[tab:params]{Hyperparameter Optimization}{This table lists the hyperparameters of OpenMax, EVM and \proser{} applied to networks trained with three loss functions, optimized on the known and negative samples of the validation set.}
  \label{tab:parameters}
\end{table}

On top of the three trained networks we incorporate post-processing methods, OpenMax, EVM, and \proser{}.
These methods require hyperparameter optimization, which we perform on the validation sets of all protocols separately to ensure optimal settings.

For OpenMax, we conduct grid search optimization for the required set of hyperparameters, including tail size $\lambda$, distance multiplier $\kappa$, and $\alpha$ as suggested by \cite{bendale2016openmax}, while relying on the cosine-based distance metric \eqref{eq:openmax-distances}.
Below are the set of values that were used to trained OpenMax models which are then validated on the validation set:

\begin{itemize}
\item Tail Size: [10, 100, 250, 500, 750, 1000]
\item Distance Multiplier = [1.5, 1.7, 2.0, 2.3]
\item Alpha = [2, 3, 5, 10]
\end{itemize}

EVM \citep{rudd2017evm} has similar hyperparameters to OpenMax, which include a tail size $\lambda$ and the distance multiplier $\kappa$.
We ended up with the following hyperparameter space to train multiple EVM models:

\begin{itemize}
\item Tail Size = [10, 25, 50, 75, 100, 150, 200, 300, 500, 1000]
\item Distance Multiplier = [0.10, 0.20, 0.30, 0.40, 0.50, 0.70, 0.90, 1.00]
\end{itemize}

Due to extensive training time, for \proser{} we use the default values for $\beta$, $\alpha$, $\lambda$ from the original implementation.
A more tedious hyperparameter optimization is left to the reader.
The only hyperparameter which we vary in our case is the number of dummy classifiers ($B$).
Since \citep{zhou2021placeholders} only trained on smaller datasets, we adapted the counts to our protocols and included:
\begin{itemize}
  \item Dummy Classifiers = [1, 2, 5, 10, 25, 100]
\end{itemize}

By using CCR@FPR criterion from \eqref{eq:ccr@fpr} for all hyperparameter combinations derived from the above list, we found the best performing hyperparameter values for both OpenMax, EVM, and PROSER in conjunction with three loss functions.
In Table~\ref{tab:parameters}, we present our findings for all three protocols, while the detailed results for the hyperparameter optimization can be found in the appendix.
For most of the hyperparameters, there seems to be no clear trend on optimality per protocol or training loss, which indicates that tuning is required for every new dataset, protocol or practical application.
The distance multipliers $\kappa$ follow opposed trends for OpenMax and EVM, since for OpenMax, original distances are computed between training samples from the same class and need to be raised, while for EVM distances between different class' training samples need to be reduced, in order to fit to the validation and test data distributions.

\begin{figure}[p]
  \subfloat[\label{fig:negatives}Negatives]{\includegraphics[width=\textwidth, page=2]{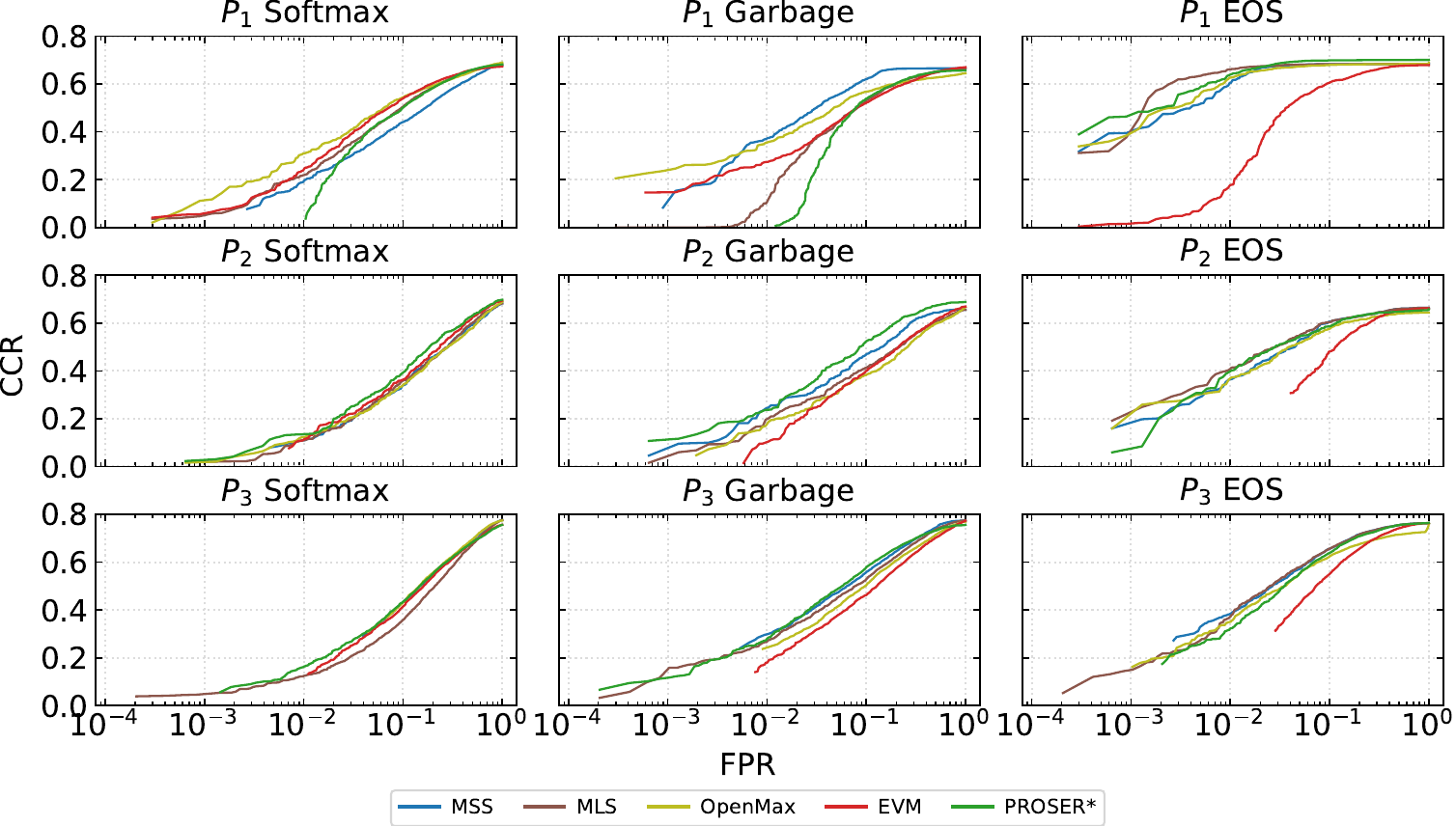}}

  \subfloat[\label{fig:unknowns}Unknowns]{\includegraphics[width=\textwidth, page=3]{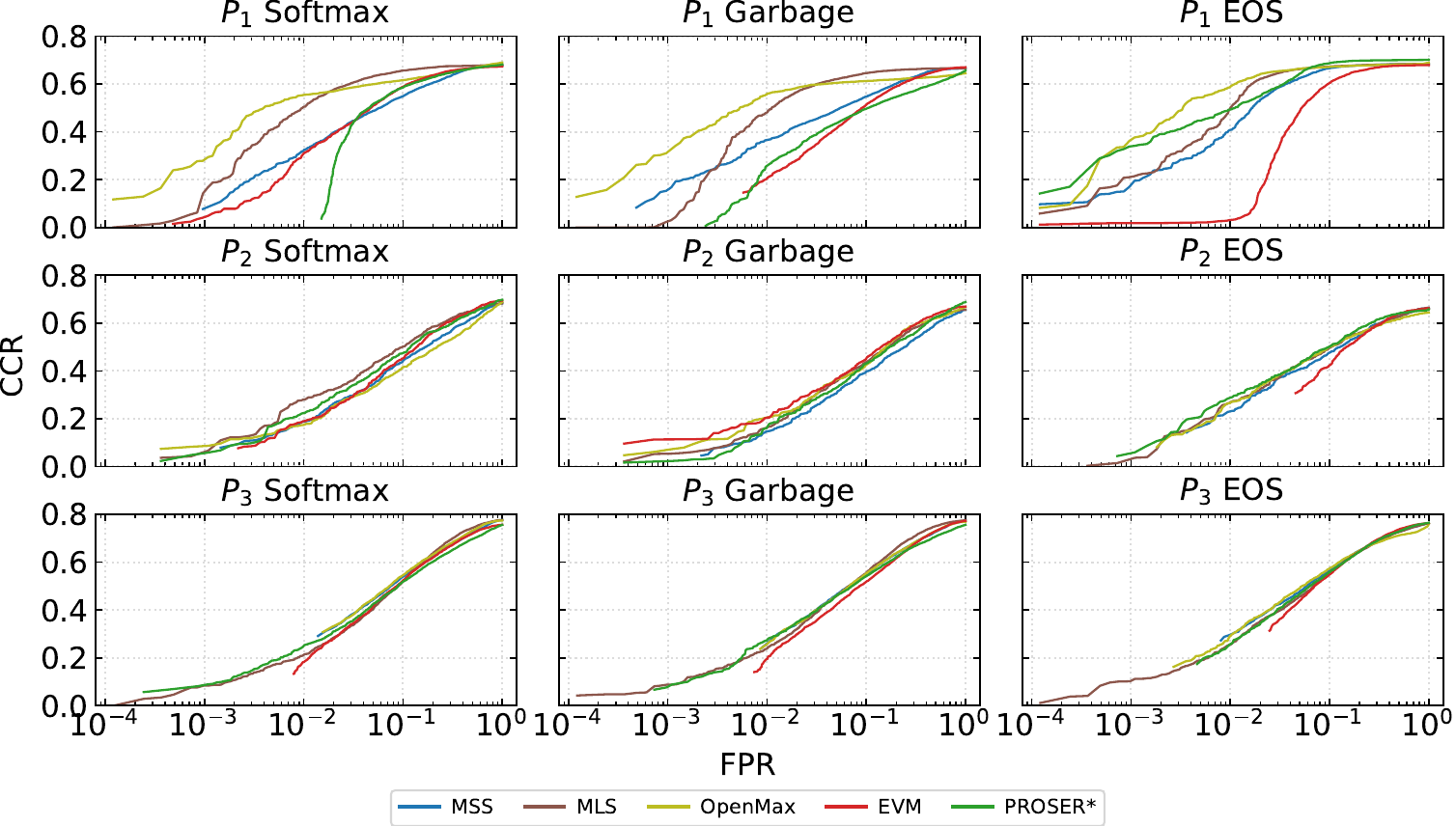}}

  \Caption[fig:OSCR]{OSCR Plots}{This figure shows OSCR plots for negative and unknown test samples on all three protocols, split across training-based methods. Colors separate post-processing methods.}
\end{figure}

\begin{table}[p]
  \setlength\tabcolsep{3pt}
  \centering\scriptsize\vspace*{-5.4ex}
  \subfloat[Protocol 1]{
    \begin{tabular}{|c|c||c||c|c|c||c||c|c|c||c|}
      \hline
      \multirow{2}{*}{\bf Post-pr.} & \multirow{2}{*}{\bf Training} & \multicolumn{4}{c||}{\bf Negative} & \multicolumn{4}{c||}{\bf Unknown} & \bf Acc \\\cline{3-11}
 & & \bf AUROC & $10^{-3}$ & $10^{-2}$ & $10^{-1}$ & \bf AUROC & $10^{-3}$ & $10^{-2}$ & $10^{-1}$ & $1$\\\hline\hline
\multirow{3}{*}{MSS} & Softmax & 0.7811 & & 0.1905 & 0.4409 & 0.8806 & 0.0772 & 0.3193 & 0.5495 & 0.6784\\
 & Garbage & \underline{0.9566} & 0.0831 & \underline{0.3710} & \underline{0.6193} & 0.8883 & 0.1533 & 0.3652 & 0.5471 & 0.6662\\
 & EOS & \it 0.9927 & \it 0.3948 & \it 0.6047 & \it 0.6836 & \it 0.9656 & \it 0.1703 & \it 0.4076 & \it 0.6666 & \it 0.6840\\
\hline
\multirow{3}{*}{MLS} & Softmax & 0.8596 & 0.0471 & 0.2186 & 0.5047 & \underline{0.9693} & 0.1422 & \it 0.5031 & \underline{0.6564} & 0.6784\\
 & Garbage & 0.8768 & 0.0000 & 0.1014 & 0.5300 & \underline{0.9676} & 0.0231 & 0.4803 & \underline{0.6459} & 0.6662\\
 & EOS & \textcolor{blue}{\bf 0.9962} & \it 0.3766 & \textcolor{blue}{\bf 0.6616} & \it 0.6834 & \textcolor{blue}{\bf 0.9792} & \it 0.2105 & 0.4962 & \it 0.6733 & \it 0.6840\\
\hline
\multirow{3}{*}{OpenMax} & Softmax & 0.8658 & \underline{0.1114} & \underline{0.3091} & \underline{0.5434} & 0.9201 & \underline{0.2791} & \underline{0.5538} & 0.6160 & \it 0.6898\\
 & Garbage & 0.8743 & \underline{0.2364} & 0.3543 & 0.5686 & 0.9146 & \underline{0.3107} & \underline{0.5553} & 0.6128 & 0.6450\\
 & EOS & \it 0.9638 & \it 0.3845 & \it 0.6267 & \it 0.6795 & \it 0.9544 & \textcolor{blue}{\bf 0.3624} & \textcolor{blue}{\bf 0.5864} & \it 0.6716 & 0.6891\\
\hline
\multirow{3}{*}{EVM} & Softmax & \underline{0.8774} & 0.0555 & 0.2381 & 0.5391 & 0.9161 & \it 0.0412 & \it 0.3067 & 0.5890 & 0.6729\\
 & Garbage & 0.8608 & \it 0.1462 & \it 0.2740 & 0.5210 & 0.8652 & & 0.2038 & 0.5131 & \underline{0.6703}\\
 & EOS & \it 0.9182 & 0.0150 & 0.1755 & \it 0.6053 & \it 0.9188 & 0.0183 & 0.0314 & \it 0.6041 & \it 0.6793\\
\hline
\multirow{3}{*}{PROSER*} & Softmax & 0.8442 & & & 0.5016 & 0.8795 & & & 0.5853 & 0.6817\\
 & Garbage & 0.8879 & & & 0.5383 & 0.8005 & & 0.2571 & 0.4967 & 0.6567\\
 & EOS & \it 0.9754 & \textcolor{blue}{\bf 0.4638} & \it 0.6384 & \textcolor{blue}{\bf 0.6990} & \it 0.9585 & \it 0.3390 & \it 0.4926 & \textcolor{blue}{\bf 0.6872} & \textcolor{blue}{\bf 0.7012}\\
\hline

    \end{tabular}
  }

  \vspace*{-1.5ex}
  \subfloat[Protocol 2]{
    \begin{tabular}{|c|c||c|c|c|c||c|c|c|c||c|}
      \hline
      \multirow{2}{*}{\bf Post-pr.} & \multirow{2}{*}{\bf Training} & \multicolumn{4}{c||}{\bf Negative} & \multicolumn{4}{c||}{\bf Unknown} & \bf Acc \\\cline{3-11}
 & & \bf AUROC & $10^{-3}$ & $10^{-2}$ & $10^{-1}$ & \bf AUROC & $10^{-3}$ & $10^{-2}$ & $10^{-1}$ & $1$\\\hline\hline
\multirow{3}{*}{MSS} & Softmax & 0.7024 & & 0.1100 & 0.3387 & 0.7726 & & 0.1887 & 0.4400 & \it 0.6827\\
 & Garbage & 0.8645 & 0.0460 & \underline{0.2427} & 0.4680 & 0.7624 & & 0.1467 & 0.3967 & 0.6560\\
 & EOS & \it 0.9181 & \it 0.1600 & \it 0.3640 & \textcolor{blue}{\bf 0.6060} & \it 0.8180 & & \it 0.2307 & \it 0.4760 & \underline{0.6640}\\
\hline
\multirow{3}{*}{MLS} & Softmax & 0.7315 & 0.0200 & 0.1087 & 0.3593 & \textcolor{blue}{\bf 0.8517} & 0.0427 & \it 0.2800 & \it 0.5013 & \it 0.6827\\
 & Garbage & 0.7828 & 0.0160 & 0.1960 & 0.4133 & 0.8210 & \it 0.0527 & 0.1613 & 0.4367 & 0.6560\\
 & EOS & \textcolor{blue}{\bf 0.9244} & \textcolor{blue}{\bf 0.1920} & \textcolor{blue}{\bf 0.4040} & \it 0.6040 & 0.8292 & 0.0147 & 0.2660 & 0.4967 & \underline{0.6640}\\
\hline
\multirow{3}{*}{OpenMax} & Softmax & 0.7301 & 0.0160 & 0.1233 & 0.3373 & 0.7589 & \it 0.0827 & 0.1733 & 0.4160 & \it 0.6880\\
 & Garbage & 0.7321 & & 0.1700 & 0.3827 & 0.7976 & 0.0613 & \underline{0.2047} & 0.4253 & 0.6673\\
 & EOS & \it 0.9119 & \it 0.1593 & \it 0.3647 & \it 0.5753 & \it 0.8268 & & \it 0.2647 & \it 0.5013 & 0.6440\\
\hline
\multirow{3}{*}{EVM} & Softmax & 0.7461 & & \it 0.1093 & 0.3627 & 0.8199 & & 0.1893 & \it 0.4520 & \it 0.6940\\
 & Garbage & 0.7732 & & 0.1007 & 0.4000 & \it 0.8282 & \textcolor{blue}{\bf 0.1133} & \it 0.2033 & \underline{0.4493} & 0.6693\\
 & EOS & \it 0.8363 & & & \it 0.4807 & 0.7920 & & & 0.4233 & 0.6620\\
\hline
\multirow{3}{*}{PROSER*} & Softmax & \underline{0.7577} & \underline{0.0233} & \underline{0.1347} & \underline{0.3913} & 0.7978 & \it 0.0480 & 0.2233 & 0.4740 & \textcolor{blue}{\bf 0.6980}\\
 & Garbage & \underline{0.8754} & \it 0.1073 & 0.2373 & \underline{0.5233} & 0.7699 & 0.0167 & 0.1600 & 0.4307 & \underline{0.6880}\\
 & EOS & \it 0.8896 & 0.0593 & \it 0.3953 & \it 0.5867 & \it 0.8302 & \underline{0.0440} & \textcolor{blue}{\bf 0.2873} & \textcolor{blue}{\bf 0.5020} & 0.6573\\
\hline

    \end{tabular}
  }

  \vspace*{-1.5ex}
  \subfloat[Protocol 3]{
    \begin{tabular}{|c|c||c|c|c|c||c|c|c|c||c|}
      \hline
      \multirow{2}{*}{\bf Post-pr.} & \multirow{2}{*}{\bf Training} & \multicolumn{4}{c||}{\bf Negative} & \multicolumn{4}{c||}{\bf Unknown} & \bf Acc \\\cline{3-11}
 & & \bf AUROC & $10^{-3}$ & $10^{-2}$ & $10^{-1}$ & \bf AUROC & $10^{-3}$ & $10^{-2}$ & $10^{-1}$ & $1$\\\hline\hline
\multirow{3}{*}{MSS} & Softmax & 0.7361 & & & 0.4331 & 0.8008 & & & 0.5399 & \textcolor{blue}{\bf 0.7768}\\
 & Garbage & 0.8389 & & \underline{0.2963} & 0.5584 & 0.8101 & & 0.2407 & 0.5493 & \underline{0.7751}\\
 & EOS & \it 0.9116 & & \textcolor{blue}{\bf 0.3826} & \it 0.6558 & \it 0.8342 & & \textcolor{blue}{\bf 0.2942} & \it 0.5681 & 0.7633\\
\hline
\multirow{3}{*}{MLS} & Softmax & 0.7202 & \underline{0.0458} & 0.1228 & 0.3587 & \underline{0.8360} & 0.0837 & 0.2106 & 0.5191 & \textcolor{blue}{\bf 0.7768}\\
 & Garbage & 0.8231 & \underline{0.1158} & 0.2661 & 0.5286 & \textcolor{blue}{\bf 0.8487} & \underline{0.0877} & 0.2380 & \underline{0.5558} & \underline{0.7751}\\
 & EOS & \textcolor{blue}{\bf 0.9146} & \textcolor{blue}{\bf 0.1437} & \it 0.3698 & \textcolor{blue}{\bf 0.6567} & 0.8341 & \textcolor{blue}{\bf 0.1019} & \it 0.2552 & \it 0.5575 & 0.7633\\
\hline
\multirow{3}{*}{OpenMax} & Softmax & 0.7400 & & & 0.4322 & 0.8104 & & & \underline{0.5460} & \it 0.7744\\
 & Garbage & 0.7799 & & 0.2375 & 0.5034 & 0.8130 & & 0.2375 & 0.5532 & 0.7694\\
 & EOS & \it 0.8292 & & \it 0.3506 & \it 0.6236 & \it 0.8147 & & \it 0.2914 & \textcolor{blue}{\bf 0.5738} & 0.7604\\
\hline
\multirow{3}{*}{EVM} & Softmax & 0.7433 & & & 0.4150 & 0.8151 & & 0.1819 & 0.5241 & 0.7567\\
 & Garbage & 0.7707 & & \it 0.1800 & 0.4628 & 0.8166 & & \it 0.1951 & 0.5193 & \it 0.7731\\
 & EOS & \it 0.8375 & & & \it 0.5498 & \it 0.8276 & & & \it 0.5482 & 0.7636\\
\hline
\multirow{3}{*}{PROSER*} & Softmax & \underline{0.7569} & & \underline{0.1612} & \underline{0.4336} & 0.7945 & \it 0.0861 & \underline{0.2462} & 0.5180 & 0.7562\\
 & Garbage & \underline{0.8456} & \it 0.1123 & 0.2766 & \underline{0.5784} & 0.7971 & 0.0766 & \it 0.2728 & 0.5416 & 0.7560\\
 & EOS & \it 0.8921 & & \it 0.3209 & \it 0.6395 & \it 0.8304 & & 0.2546 & \it 0.5575 & \it 0.7649\\
\hline

    \end{tabular}
  }
  \vspace*{-2ex}
  \Caption[tab:CCR]{Detailed Results}{This table shows detailed AUROC and CCR values on Negative and Unknown test samples for several FPR thresholds $\zeta$, as well as the closed-set accuracy. In each column, the best values are highlighted in \textcolor{blue}{\bf blue}, the best per post-processing method in \textit{italics}, and the best per training loss are \underline{underlined}. When low FPR values cannot be reached with any threshold, fields are left empty.}
  \end{table}

\section{Results and Observations}
\label{sec:discussion}

Utilizing the optimized hyperparameters of the post-processing methods, we switch our evaluation to the test set.
\fig{OSCR} shows OSCR curves with logarithmic scale for FPR axes for the 15 experimental settings on three protocols.
\sfig{negatives} here illustrates the test set performance for the negative, and \sfig{unknowns} for the unknown test samples.
Each of them is split into the three protocols and the three training-based methods, while different colors indicate different post-processing methods.
A more compact but also much more busy plot is provided in the appendix.
Additionally, we list AUROC scores as well as CCR values for selected FPR values in \tab{CCR}, for both negative and unknown test samples.
Within each protocol, we observe that majority of the classifiers show similar CCR performances in the closed-set case where FPR=1.

%It is evident from the plots that scenarios involving EOS mostly outperform the other two training methods for both negative and unknown samples.
%Indeed, EOS alone always appears in top three for negative cases across all protocols.
%Besides EOS, OpenMax also seems to enhance the performance when it is combined with other training methods.
%On the other hand, EVM-involved cases usually perform around the baseline or worse, especially for the unknowns in protocol \p1.
%For the same protocol, we see that MLS performs well for both negative and unknown samples.
%Indeed, MLS coupled with EOS shows an outstanding performance when tested on negative samples.
%In the following protocols, EOS+MLS still appears among the well performing scenarios for negative samples with a slight deterioration.

%Similar to EOS, Garbage training seems to help improve performances of algorithms when combined, but experiences a similar performance loss when combined with EVM.
%However, it still shows a better performance in discriminating negative and unknown samples by itself with respect to SoftMax in all protocols.
%Finally, we tend to see an overall drop in performances when it comes to unknown samples in protocols \p2 and \p3, and observe very close OSCR curves.

Since we provide many results in \fig{OSCR} and \tab{CCR}, we cannot discuss every single combination of methods separately.
This section provides some highlights on the main conclusions that can be drawn from our experiments.

\subsection{Negative vs. Unknown Test Samples}
In our experiments, we evaluated our results with both negative and unknown test samples to compute CCR and FPR values -- as given in \tab{CCR} -- and OSCR plots provided in \fig{OSCR}.
For all protocols, we had selected the negative samples to be semantically closer to the known samples than the unknowns, cf.~\fig{protocols}, except for protocol \p3 where knowns, negatives and unknowns are semantically similar.
Consequently, when ignoring the negatives during training as done with the SoftMax approach (left column in \fig{OSCR}), results on the unknown set are actually higher than on the negative set.
Also for \proser{} (green lines in \fig{OSCR}), which samples negatives from mixing known samples instead, a similar behavior can be observed.

Observations change completely when using algorithms that make use of our provided negative samples during training.
Please remember that the negative \emph{sample instances} used during training and validation differ from the ones used for testing -- only the classes are shared.
Particularly, training with Entropic Open-Set loss (right column in \fig{OSCR}) improves extremely on the evaluation of negative test samples, but also the Garbage class approach (middle column in \fig{OSCR}) can benefit.
Hence, when we know which kinds of uninteresting samples will most likely be seen during testing, these types of samples shall be used as negatives in combination with EOS to train the network.
However, when we cannot know which kinds of unknown samples to expect, different solutions need to be found, for example, using manifold mix-up \citep{verma2019manifold} or adversarial samples \citep{chen2022reciprocal} as negatives.
In this case, additional open-set post-processing methods on top of these deep features, for example OpenMax (tan lines in \fig{OSCR}), might be used to improve performance.

\subsection{OSCR vs. AUROC}
\label{sec:OSCR-AUROC}
For comparison, we also added AUROC values to \tab{CCR}.
We can observe that all maximal AUROC values (highlighted in blue) appear in the MLS algorithm, combined with various training methods.
This seems to validate the claim of \cite{vaze2022openset} that MLS performs very well in any open-set task and that no other open-set algorithm is able to outperform MLS.
However, when having a closer look into the CCR values at various FPR thresholds, especially when evaluating unknown test samples, it is apparent that such a conclusion is misleading.
Good results can be observed for various combinations of training and post-processing algorithms, and depending on the selected FPR threshold and the difficulty of the evaluation protocol, conclusions can differ tremendously.

\subsection{Easy vs. Hard Unknowns}
One particular advantage of our proposed evaluation protocols is the possibility to evaluate open-set scenarios in different difficulties.
Indeed, when looking into the results on the different protocols, we can see major differences in the advances of open-set evaluations.
When unknown classes are very different from the known classes, as given in protocol \p1, post-processing approaches such as OpenMax or MLS can bring large improvements over the standard thresholding of SoftMax scores (MSS).
Particularly, the combination of training the network with EOS and using OpenMax on top of these deep features seems to be the most fruitful approach here.

This changes drastically when evaluating the more difficult open-set protocol \p3, where MLS actually decreases performance with respect to the simple MSS approach.
Also, OpenMax cannot really improve over the MSS baseline.
Only the EOS loss seems to provide slight improvements over the Garbage class approach, which in turn has shows tiny improvements over the SoftMax baseline.
However, especially when testing with unknown samples, all results are very similar and no clear winner can be identified.

\subsection{Garbage Class vs. Entropic Open-Set Loss}
In virtually none of our evaluations, the prevailing approach in the literature -- collecting all negative samples in a separate Garbage class during training -- has been the best approach.
Throughout, the Garbage approach was inferior to EOS.
Forcing all negative (and unknown) samples to reside in a separate part of the feature space, as done by Garbage (cf.~for example \cite{dhamija2018agnostophobia}), does not seem to be the best idea.
Instead, encouraging the deep features of negative (and unknown) samples to have small magnitudes enforces the natural behavior of deep networks, and is proven by \cite{dhamija2018agnostophobia} to be an optimal way of restricting confidence scores of negative samples.

However, there were some differences in the training procedure.
For example, the Garbage approach used class-balancing weights as given in \eqref{eq:weights-bg}, which reduced the influence of negative samples on the loss -- if such class weights were not present, only the majority class (the negative class in our case) would have been learned well \citep{rudd2016moon}.
On the other hand, for EOS training, we kept the weights of each negative sample as high as for the known samples, which made the training focus more on negative samples.
Especially in protocols \p2 and \p3, where known and negative samples are extremely similar, this lead to a noticeable decrease in closed-set performance.
In future work, we will investigate whether a class weighting scheme between the two extremes can help in keeping high closed-set accuracy while still improving open-set capabilities.

\subsection{EVM and OpenMax}
Both EVM and OpenMax estimate Weibull distributions on deep feature representations for estimating the probability of sample inclusion.
Only the data used to estimate these distributions differ.
While OpenMax only looks into the distribution of deep feature distances inside classes, EVM models such distances between classes.
Hence, EVM requires the distribution of different-class known samples to appropriately predict the distribution of unknown samples, which might be too difficult in general.
EVM hyperparameters optimized on the negative samples of the validation data are likely to translate badly to unknown samples from test data.
Since the distribution of negative and unknown classes purposefully differs, especially in protocol \p1, EVM (red lines in \fig{OSCR}) is highly inferior to MSS (blue lines in \fig{OSCR}).
However, even when negative and unknown data is similar as in \p3, EVM is not able to show improvements.

On the other hand, OpenMax depends only on same-class feature comparisons, which makes it independent of the distribution of different-class and unknown samples.
Therefore, it can handle well unknown samples that are far away from the knowns, as in protocol \p1.
However, it loses its advantage in more complex situations where known and unknown data are more semantically similar.

Note that we had to make use of the distance multiplier $\kappa$ in \eqref{eq:openmax-distances} to model samples correctly.
We assume that this problem arises since we model activation vectors from the same data that the network is trained with, which might not follow the distribution of the test data.
In future work, we will investigate whether using validation data for training OpenMax and EVM will improve the similarity between model training and test data.
Also, theoretically we could add the negative data to our EVM as additional other-class samples, which might improve its stability, at least when evaluated on negative data.

\subsection{\proser}
For protocols \p1 and \p2, \proser{} provides the best performances at large FPR thresholds 0.1 and 1, and reasonable CCR scores at smaller FPR thresholds.
Especially when the network is originally trained with EOS loss, \proser{} achieves the best results in our evaluation.
Presumably, EOS, which makes use of real negative samples during training, shapes the feature space such that negative samples cluster in the center of the feature space.
This provides a good initialization for \proser{}, which utilizes mix-up negative samples, which further shape the space for unknown samples.

In any case, these results do not transfer to more difficult cases, as provided in protocol \p3, where \proser{} fine-tuning is actually decremental and results are generally reduced w.r.t.~the MSS or MLS baseline.

%However, we have heard from other sources that PROSER does not work well in large-scale evaluations, so we are not surprised by our results.
%For this reason, it is important to have more realistic and large-scale evaluations of open-set techniques to show their behavior in practice, and do not rely on small-scale and simple evaluations as provided in most research papers on open-set classification.

%Note, however, that parameter optimization in PROSER always requires a full network training, for which we did not have the resources to do.

\section{Combining Training-based with Post-processing Methods}

\begin{figure}[t]
  \centering
  \includegraphics[width=.95\textwidth, page=4]{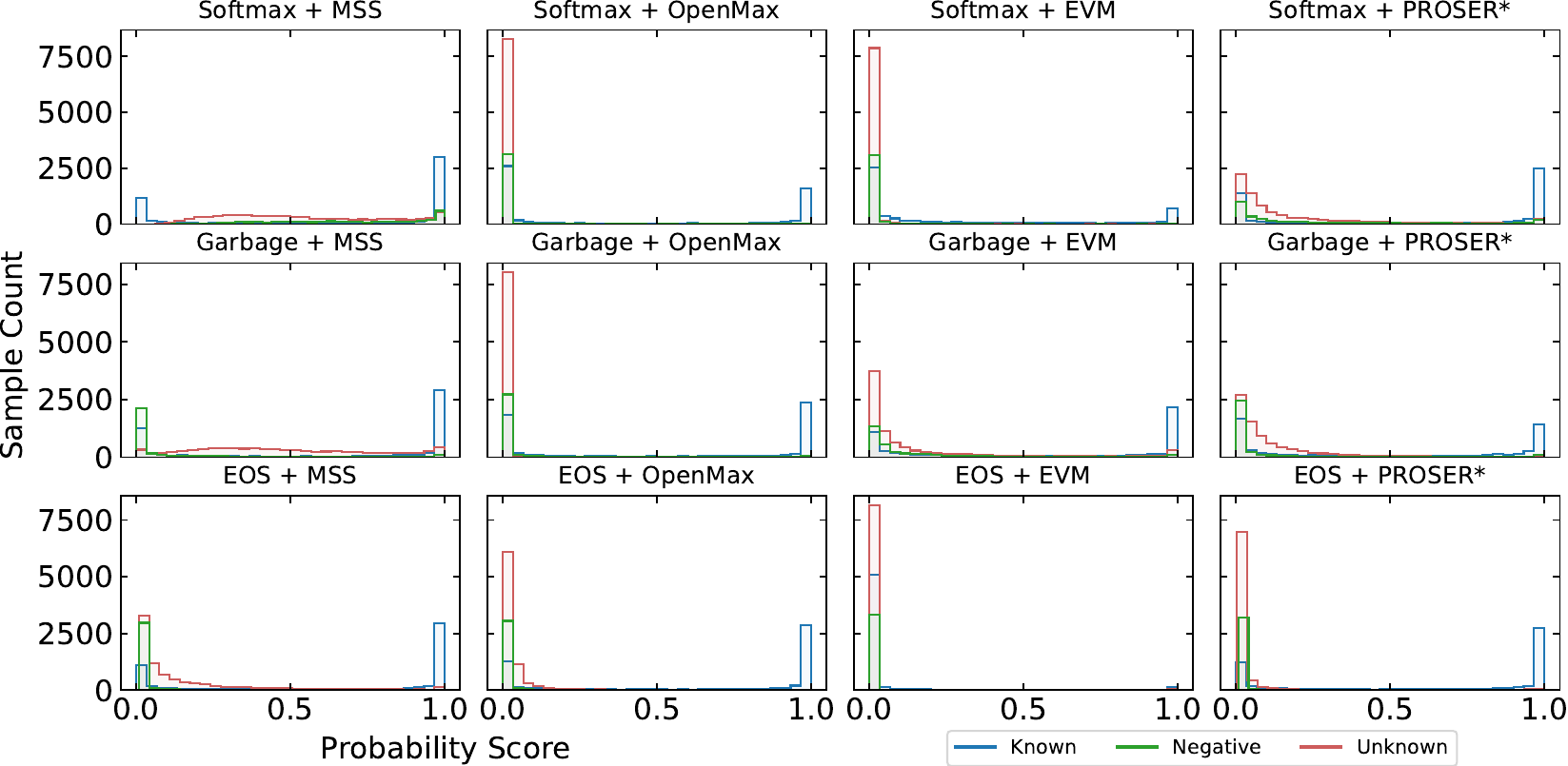}
  \Caption[fig:distributions]{Score Distributions}{This figure shows score distributions extracted from the network trained with three different loss functions, and further post-processed with all algorithms (except for MLS), on protocol \p1.}
\end{figure}

In this paper, for the first time in research, two different basic approaches to open-set classification are combined, which are training-based methods such as EOS, and post-processing methods such as OpenMax or \proser.
In order to show the advantages of this combination, we plot the distributions of the probability scores extracted from different networks obtained through different algorithms.
Here we exclude MLS since it does not provide probability distributions, but only logit scores.
Particularly, for known samples, we use the probability assigned to the correct class, while for negative and unknown samples, we select the maximal probability obtained over all known classes.
The score distributions for protocol \p1 can be found in \fig{distributions}, while for the other two protocols they are in the appendix.
The discussion here mainly targets protocol \p1 in \fig{distributions}, but similar conclusions can be drawn from the other two protocols.
Interestingly, we can observe an almost binary score distribution, where probabilities are either close to 0 or close to 1.
This validates the well-known fact \cite{matan1990handwritten} that SoftMax scores are poor probability estimates.

When training a network with the SoftMax approach, the network learns to provide large confidence scores for known samples, but also assigns larger scores to negative and unknown samples.
When trying to dissect these using OpenMax, EVM, or \proser, most of the negative and unknown samples get very low maximal scores, but also many of the known samples reduce their confidences, usually drastically.
Training the network with a Garbage class does not improve score distributions when thresholding probabilities, but OpenMax and EVM can keep high probability scores for known samples -- yet a few unknown samples still obtain a probability close to 1 of being classified as a known class.
Finally, EOS provides a feature space where most negative and unknown samples have low confidence scores, and OpenMax and \proser{} can further reduce the number of unknown samples that have high scores while not reducing scores for known classes -- EVM on the other hand seems not to profit from better separation in the original feature space.

\iffalse
\subsection{Evaluation Metric}
\td{I am not sure if we should have this subsection inside. Please remove if you believe that this would weaken our paper.}
In this evaluation, we made use of the OSCR curve for evaluating open-set classification, which is an adoption of the Open-ROC curve used in evaluating open-set face recognition.
We adopted this metric since it was used in the original paper \cite{palechor2023protocols} and in \cite{dhamija2018agnostophobia}, and it is now widely used in other evaluations \cite{vaze2022openset}.
While in face recognition, a larger focus is laid on reducing the number of false positives, this might not reflect the use case in open-set image classification.
In future work, metrics that reflect better the use cases in applied open-set classification need to be developed and used in this evaluation.

Luckily, we provide the source code for the entire paper on GitHub\footref{fn:package}, so that researchers can vary the evaluation metric toward their needs, without changing the underlying methods.
We are currently also implementing other published open-set training techniques into this repository, to finally provide good insights into which directions open-set classification might need to go to.
\fi

\section{Conclusion}
\label{sec:conclusion}

The aim of this paper is to provide the first large-scale and reproducible evaluation of different open-set classification techniques.
Based on the ILSVRC 2012 data set and the WordNet hierarchy, we developed three open-set evaluation protocols that contain different difficulties in terms of closed-set and open-set classification.
We experimented with three different loss functions to train initial networks to obtain deep features that are suited for open-set classification.
On top of these networks, we applied five different post-processing techniques.
We showed that both better features and additional open-set techniques can improve open-set performance.
For the first time in the literature, we also combined feature learning and post-processing techniques.
In the cases that we know which kinds of unknown samples can be expected, and when known and unknown classes are semantically very different, the combination of Entropic Open-Set loss to obtain separable features and post-processing with OpenMax or \proser{} to reject unknown samples provided the best improvements.
On the other hand, when known and unknown classes are semantically similar, none of the evaluated techniques provided reasonable improvements over thresholding standard SoftMax confidence scores (MSS).

In this evaluation we only made use of a single network topology -- ResNet-50 with an additional embedding layer.
Future research should investigate whether other network topologies follow the same trends.
Also, other open-set classification techniques have been proposed in the literature, which need to be tested in large scale as well.
For example, in our protocols we solely made use of real negative samples -- except during \proser{} post-processing -- while the literature is largely focused on how to artificially create such negative samples.
Including different negative sample generation techniques will open a third type of algorithm, which can be combined with the currently implemented two types.
Since we provide our complete source code -- from raw images to network training, open-set modeling and the final plots and tables in this paper and its appendix -- testing your own algorithm in a realistic, comparable and reproducible manner is just one step away!

\newpage
\bibliography{References,Publications,Theses}

\begin{thebibliography}{44}
\providecommand{\natexlab}[1]{#1}
\providecommand{\url}[1]{\texttt{#1}}
\expandafter\ifx\csname urlstyle\endcsname\relax
  \providecommand{\doi}[1]{doi: #1}\else
  \providecommand{\doi}{doi: \begingroup \urlstyle{rm}\Url}\fi

\bibitem[Bendale and Boult(2016)]{bendale2016openmax}
Abhijit Bendale and Terrance~E. Boult.
\newblock Towards open set deep networks.
\newblock In \emph{Conference on Computer Vision and Pattern Recognition
  (CVPR)}. IEEE, 2016.

\bibitem[Boult et~al.(2019)Boult, Cruz, Dhamija, G\"unther, Henrydoss, and
  Scheirer]{boult2019learning}
Terrance~E. Boult, Steve Cruz, Akshay~Raj Dhamija, Manuel G\"unther, James
  Henrydoss, and Walter~J. Scheirer.
\newblock Learning and the unknown: Surveying steps toward open world
  recognition.
\newblock In \emph{AAAI Conference on Artificial Intelligence}, volume~33,
  pages 9801--9807, 2019.

\bibitem[Chen et~al.(2020)Chen, Qiao, Shi, Peng, Li, Huang, Pu, and
  Tian]{chen2020learning}
Guangyao Chen, Limeng Qiao, Yemin Shi, Peixi Peng, Jia Li, Tiejun Huang,
  Shiliang Pu, and Yonghong Tian.
\newblock Learning open set network with discriminative reciprocal points.
\newblock In \emph{European Conference on Computer Vision (ECCV)}. Springer,
  2020.

\bibitem[Chen et~al.(2022)Chen, Peng, Wang, and Tian]{chen2022reciprocal}
Guangyao Chen, Peixi Peng, Xiangqian Wang, and Yonghong Tian.
\newblock Adversarial reciprocal points learning for open set recognition.
\newblock \emph{Transactions on Pattern Analysis and Machine Intelligence
  (TPAMI)}, 44\penalty0 (11), 2022.

\bibitem[Deng et~al.(2009)Deng, Dong, Socher, Li, Li, and
  Fei-Fei]{deng2009imagenet}
Jia Deng, Wei Dong, Richard Socher, Li-Jia Li, Kai Li, and Li~Fei-Fei.
\newblock {ImageNet}: A large-scale hierarchical image database.
\newblock In \emph{Conference on Computer Vision and Pattern Recognition
  (CVPR)}. IEEE, 2009.

\bibitem[Dhamija et~al.(2018)Dhamija, G\"unther, and
  Boult]{dhamija2018agnostophobia}
Akshay~Raj Dhamija, Manuel G\"unther, and Terrance~E. Boult.
\newblock Reducing network agnostophobia.
\newblock In \emph{Advances in Neural Information Processing Systems
  (NeurIPS)}, pages 9157--9168, 2018.

\bibitem[Dhamija et~al.(2019)Dhamija, G\"unther, and
  Boult]{dhamija2019improving}
Akshay~Raj Dhamija, Manuel G\"unther, and Terrance~E. Boult.
\newblock Improving deep network robustness to unknown inputs with
  {Objectosphere}.
\newblock In \emph{Conference on Computer Vision and Patter Recognition
  Workshops (CVPRW)}, 2019.

\bibitem[Dhamija et~al.(2020)Dhamija, G\"unther, Ventura, and
  Boult]{dhamija2020elephant}
Akshay~Raj Dhamija, Manuel G\"unther, Jonathan Ventura, and Terrance~E. Boult.
\newblock The overlooked elephant of object detection: Open set.
\newblock In \emph{Winter Conference on Applications of Computer Vision
  (WACV)}, pages 1021--1030, 2020.

\bibitem[Ge et~al.(2017)Ge, Demyanov, and Garnavi]{ge2017generative}
Zongyuan Ge, Sergey Demyanov, and Rahil Garnavi.
\newblock Generative {OpenMax} for multi-class open set classification.
\newblock In \emph{British Machine Vision Conference (BMVC)}, 2017.

\bibitem[Geng et~al.(2021)Geng, Huang, and Chen]{geng2021recent}
Chuanxing Geng, Sheng-Jun Huang, and Songcan Chen.
\newblock Recent advances in open set recognition: A survey.
\newblock \emph{Transactions on Pattern Analysis and Machine Intelligence
  (TPAMI)}, 43\penalty0 (10):\penalty0 3614--3631, 2021.

\bibitem[Grother et~al.(2022)Grother, Ngan, and Hanaoka]{nist2022ongoing}
Patrick Grother, Mei Ngan, and Kayee Hanaoka.
\newblock Face recognition vendor test ({FRVT}) part 2: Identification.
\newblock Technical report, National Institute of Standards and Technology,
  2022.

\bibitem[G\"unther et~al.(2017)G\"unther, Cruz, Rudd, and
  Boult]{guenther2017toward}
Manuel G\"unther, Steve Cruz, Ethan~M. Rudd, and Terrance~E. Boult.
\newblock Toward open-set face recognition.
\newblock In \emph{Conference on Computer Vision and Pattern Recognition
  Workshops (CVPRW)}, pages 71--80, 2017.

\bibitem[G\"unther et~al.(2020)G\"unther, Dhamija, and
  Boult]{guenther2020watchlist}
Manuel G\"unther, Akshay~Raj Dhamija, and Terrance~E. Boult.
\newblock Watchlist adaptation: Protecting the innocent.
\newblock In \emph{International Conference of the Biometrics Special Interest
  Group (BIOSIG)}, 2020.

\bibitem[He et~al.(2016)He, Zhang, Ren, and Sun]{he2016deep}
Kaiming He, Xiangyu Zhang, Shaoqing Ren, and Jian Sun.
\newblock Deep residual learning for image recognition.
\newblock In \emph{Conference on Computer Vision and Pattern Recognition
  (CVPR)}. IEEE, 2016.

\bibitem[Hendrycks and Gimpel(2017)]{hendrycks2017baseline}
Dan Hendrycks and Kevin Gimpel.
\newblock A baseline for detecting misclassified and out-of-distribution
  examples in neural networks.
\newblock In \emph{International Conference on Learning Representations
  (ICLR)}, 2017.

\bibitem[Hendrycks et~al.(2022)Hendrycks, Basart, Mazeika, Zou, Kwon,
  Mostajabi, Steinhardt, and Song]{hendrycks2022scaling}
Dan Hendrycks, Steven Basart, Mantas Mazeika, Andy Zou, Joseph Kwon,
  Mohammadreza Mostajabi, Jacob Steinhardt, and Dawn Song.
\newblock Scaling out-of-distribution detection for real-world settings.
\newblock In \emph{International Conference on Machine Learning (ICML)}. PMLR,
  2022.

\bibitem[Huang et~al.(2017)Huang, Liu, van~der Maaten, and
  Weinberger]{huang2017densenet}
Gao Huang, Zhuang Liu, Laurens van~der Maaten, and Kilian~Q. Weinberger.
\newblock Densely connected convolutional networks.
\newblock In \emph{Conference on Computer Vision and Pattern Recognition
  (CVPR)}, 2017.

\bibitem[Jiang and Learned-Miller(2017)]{jiang2017face}
Huaizu Jiang and Erik Learned-Miller.
\newblock Face detection with the {Faster R-CNN}.
\newblock In \emph{International Conference on Automatic Face \& Gesture
  Recognition (FG)}, 2017.

\bibitem[Krizhevsky and Hinton(2009)]{krizhevsky2009cifar}
Alex Krizhevsky and Geoffrey Hinton.
\newblock Learning multiple layers of features from tiny images.
\newblock Technical report, University of Toronto, 2009.

\bibitem[Krizhevsky et~al.(2012)Krizhevsky, Sutskever, and
  Hinton]{krizhevsky2012alexnet}
Alex Krizhevsky, Ilya Sutskever, and Geoffrey~E. Hinton.
\newblock {ImageNet} classification with deep convolutional neural networks.
\newblock In \emph{Advances in Neural Information Processing Systems (NIPS)},
  2012.

\bibitem[Le and Yang(2015)]{le2015tiny}
Ya~Le and Xuan Yang.
\newblock Tiny imagenet visual recognition challenge.
\newblock Technical report, Stanford, 2015.

\bibitem[LeCun et~al.(1995)LeCun, Jackel, Bottou, Cortes, Denker, Drucker,
  Guyon, Muller, Sackinger, Simard, et~al.]{lecun1995learning}
Yann LeCun, LD~Jackel, L{\'e}on Bottou, Corinna Cortes, John~S Denker, Harris
  Drucker, Isabelle Guyon, UA~Muller, E~Sackinger, Patrice Simard, et~al.
\newblock Learning algorithms for classification: A comparison on handwritten
  digit recognition.
\newblock \emph{Neural networks: the statistical mechanics perspective},
  261:\penalty0 276, 1995.

\bibitem[LeCun et~al.(1998)LeCun, Cortes, and Burges]{lecun1998mnist}
Yann LeCun, Corinna Cortes, and Christopher J.~C. Burges.
\newblock The {MNIST} database of handwritten digits, 1998.

\bibitem[Liang et~al.(2017)Liang, Li, and Srikant]{liang2017odin}
Shiyu Liang, Yixuan Li, and Rayadurgam Srikant.
\newblock Enhancing the reliability of out-of-distribution image detection in
  neural networks.
\newblock In \emph{International Conference on Learning Representations
  (ICLR)}, 2017.

\bibitem[Matan et~al.(1990)Matan, Kiang, Stenard, Boser, Denker, Henderson,
  Howard, Hubbard, Jackel, and Le~Cun]{matan1990handwritten}
Ofer Matan, R.K. Kiang, C.E. Stenard, B.~Boser, J.S. Denker, D.~Henderson, R.E.
  Howard, W.~Hubbard, L.D. Jackel, and Yann Le~Cun.
\newblock Handwritten character recognition using neural network architectures.
\newblock In \emph{USPS Advanced Technology Conference}, 1990.

\bibitem[Miller(1998)]{miller1998wordnet}
George~A Miller.
\newblock \emph{WordNet: An electronic lexical database}.
\newblock MIT press, 1998.

\bibitem[Neal et~al.(2018)Neal, Olson, Fern, Wong, and
  Li]{neal2018counterfactual}
Lawrence Neal, Matthew Olson, Xiaoli Fern, Weng-Keen Wong, and Fuxin Li.
\newblock Open set learning with counterfactual images.
\newblock In \emph{European Conference on Computer Vision (ECCV)}, 2018.

\bibitem[Netzer et~al.(2011)Netzer, Wang, Coates, Bissacco, Wu, and
  Ng]{netzer2011svhn}
Yuval Netzer, Tao Wang, Adam Coates, Alessandro Bissacco, Bo~Wu, and Andrew~Y.
  Ng.
\newblock Reading digits in natural images with unsupervised feature learning.
\newblock In \emph{Advances in Neural Information Processing Systems (NIPS)
  Workshop}, 2011.

\bibitem[Palechor et~al.(2023)Palechor, Bhoumik, and
  G\"unther]{palechor2023protocols}
Andres Palechor, Annesha Bhoumik, and Manuel G\"unther.
\newblock Large-scale open-set classification protocols for imagenet.
\newblock In \emph{Winter Conference on Applications of Computer Vision
  (WACV)}, pages 42--51. CVF/IEEE, January 2023.

\bibitem[Phillips et~al.(2011)Phillips, Grother, and
  Micheals]{phillips2011evaluation}
P.~Jonathon Phillips, Patrick Grother, and Ross Micheals.
\newblock \emph{Handbook of Face Recognition}, chapter Evaluation Methods in
  Face Recognition.
\newblock Springer, 2nd edition, 2011.

\bibitem[Redmon et~al.(2016)Redmon, Divvala, Girshick, and
  Farhadi]{redmon2016yolo}
Joseph Redmon, Santosh Divvala, Ross Girshick, and Ali Farhadi.
\newblock You only look once: Unified, real-time object detection.
\newblock In \emph{Conference on Computer Vision and Pattern Recognition
  (CVPR)}. IEEE, 2016.

\bibitem[Ren et~al.(2015)Ren, He, Girshick, and Sun]{ren2015faster}
Shaoqing Ren, Kaiming He, Ross Girshick, and Jian Sun.
\newblock Faster {R-CNN}: Towards real-time object detection with region
  proposal networks.
\newblock In \emph{Advances in Neural Information Processing Systems (NIPS)},
  2015.

\bibitem[Rudd et~al.(2016)Rudd, G\"unther, and Boult]{rudd2016moon}
Ethan~M. Rudd, Manuel G\"unther, and Terrance~E. Boult.
\newblock {MOON}: A mixed objective optimization network for the recognition of
  facial attributes.
\newblock In \emph{European Conference on Computer Vision (ECCV)}, pages
  19--35. Springer, 2016.

\bibitem[Rudd et~al.(2017)Rudd, Jain, Scheirer, and Boult]{rudd2017evm}
Ethan~M. Rudd, Lalit~P. Jain, Walter~J. Scheirer, and Terrance~E. Boult.
\newblock The extreme value machine.
\newblock \emph{Transactions on Pattern Analysis and Machine Intelligence
  (TPAMI)}, 2017.

\bibitem[Russakovsky et~al.(2015)Russakovsky, Deng, Su, Krause, Satheesh, Ma,
  Huang, Karpathy, Khosla, Bernstein, et~al.]{russakovsky2015imagenet}
Olga Russakovsky, Jia Deng, Hao Su, Jonathan Krause, Sanjeev Satheesh, Sean Ma,
  Zhiheng Huang, Andrej Karpathy, Aditya Khosla, Michael Bernstein, et~al.
\newblock Imagenet large scale visual recognition challenge.
\newblock \emph{International Journal of Computer Vision (IJCV)}, 115\penalty0
  (3), 2015.

\bibitem[Vaze et~al.(2022)Vaze, Han, Vedaldi, and Zissermann]{vaze2022openset}
Sagar Vaze, Kai Han, Andrea Vedaldi, and Andrew Zissermann.
\newblock Open-set recognition: A good closed-set classifier is all you need?
\newblock In \emph{International Conference on Learning Representations
  (ICLR)}, 2022.

\bibitem[Verma et~al.(2019)Verma, Lamb, Beckham, Najafi, Mitliagkas, Lopez-Paz,
  and Bengio]{verma2019manifold}
Vikas Verma, Alex Lamb, Christopher Beckham, Amir Najafi, Ioannis Mitliagkas,
  David Lopez-Paz, and Yoshua Bengio.
\newblock Manifold mixup: Better representations by interpolating hidden
  states.
\newblock In \emph{International Conference on Machine Learning (ICML)}. PMLR,
  2019.

\bibitem[Wang et~al.(2022)Wang, Xu, Yang, He, Cao, and Huang]{wang2022openauc}
Zitai Wang, Qianqian Xu, Zhiyong Yang, Yuan He, Xiaochun Cao, and Qingming
  Huang.
\newblock Openauc: Towards auc-oriented open-set recognition.
\newblock In \emph{Advances in Neural Information Processing Systems
  (NeuRIPS)}, 2022.

\bibitem[Wilson et~al.(2023)Wilson, Fischer, Dayoub, Miller, and
  S{\"u}nderhauf]{wilson2023safe}
Samuel Wilson, Tobias Fischer, Feras Dayoub, Dimitry Miller, and Niko
  S{\"u}nderhauf.
\newblock {SAFE}: Sensitivity-aware features for out-of-distribution object
  detection.
\newblock \emph{arXiv preprint arXiv:2208.13930}, 2023.

\bibitem[Xiao et~al.(2017)Xiao, Rasul, and Vollgraf]{xiao2017fashion}
Han Xiao, Kashif Rasul, and Roland Vollgraf.
\newblock Fashion-{MNIST}: A novel image dataset for benchmarking machine
  learning algorithms.
\newblock \emph{arXiv preprint arXiv:1708.07747}, 2017.

\bibitem[Yang et~al.(2022)Yang, Wang, Zou, Zhou, Ding, Peng, Wang, Chen, Li,
  Sun, Du, Zhou, Zhang, Hendrycks, Li, and Liu]{yang2022openood}
Jingkang Yang, Pengyun Wang, Dejian Zou, Zitang Zhou, Kunyuan Ding, Wenxuan
  Peng, Haoqi Wang, Guangyao Chen, Bo~Li, Yiyou Sun, Xuefeng Du, Kaiyang Zhou,
  Wayne Zhang, Dan Hendrycks, Yixuan Li, and Ziwei Liu.
\newblock Openood: Benchmarking generalized out-of-distribution detection.
\newblock In \emph{Advances in Neural Information Processing Systems
  (NeuRIPS)}, 2022.

\bibitem[Yuan et~al.(2021)Yuan, Chen, Wang, Yu, Shi, Jiang, Tay, Feng, and
  Yan]{yuan2021tokens}
Li~Yuan, Yunpeng Chen, Tao Wang, Weihao Yu, Yujun Shi, Zi-Hang Jiang,
  Francis~E.H. Tay, Jiashi Feng, and Shuicheng Yan.
\newblock {Tokens-to-Token ViT}: Training vision transformers from scratch on
  imagenet.
\newblock In \emph{International Conference on Computer Vision (ICCV)}, 2021.

\bibitem[Zhao et~al.(2019)Zhao, Zheng, Xu, and Wu]{zhao2019object}
Zhong-Qiu Zhao, Peng Zheng, Shou-tao Xu, and Xindong Wu.
\newblock Object detection with deep learning: A review.
\newblock \emph{Transactions on Neural Networks and Learning Systems (TNNLS)},
  30\penalty0 (11):\penalty0 3212--3232, 2019.

\bibitem[Zhou et~al.(2021)Zhou, Ye, and Zhan]{zhou2021placeholders}
Da-Wei Zhou, Han-Jia Ye, and De-Chuan Zhan.
\newblock Learning placeholders for open-set recognition.
\newblock In \emph{Conference on Computer Vision and Pattern Recognition
  (CVPR)}, 2021.

\end{thebibliography}

\newpage
\section*{Appendix}

Here, we provide the additional score distributions plots for protocols \p2 and \p3.

\begin{figure}[H]
  \centering
  \subfloat[Protocol \p2\label{fig:distributions:P2}]{
    \includegraphics[width=.95\textwidth, page=5]{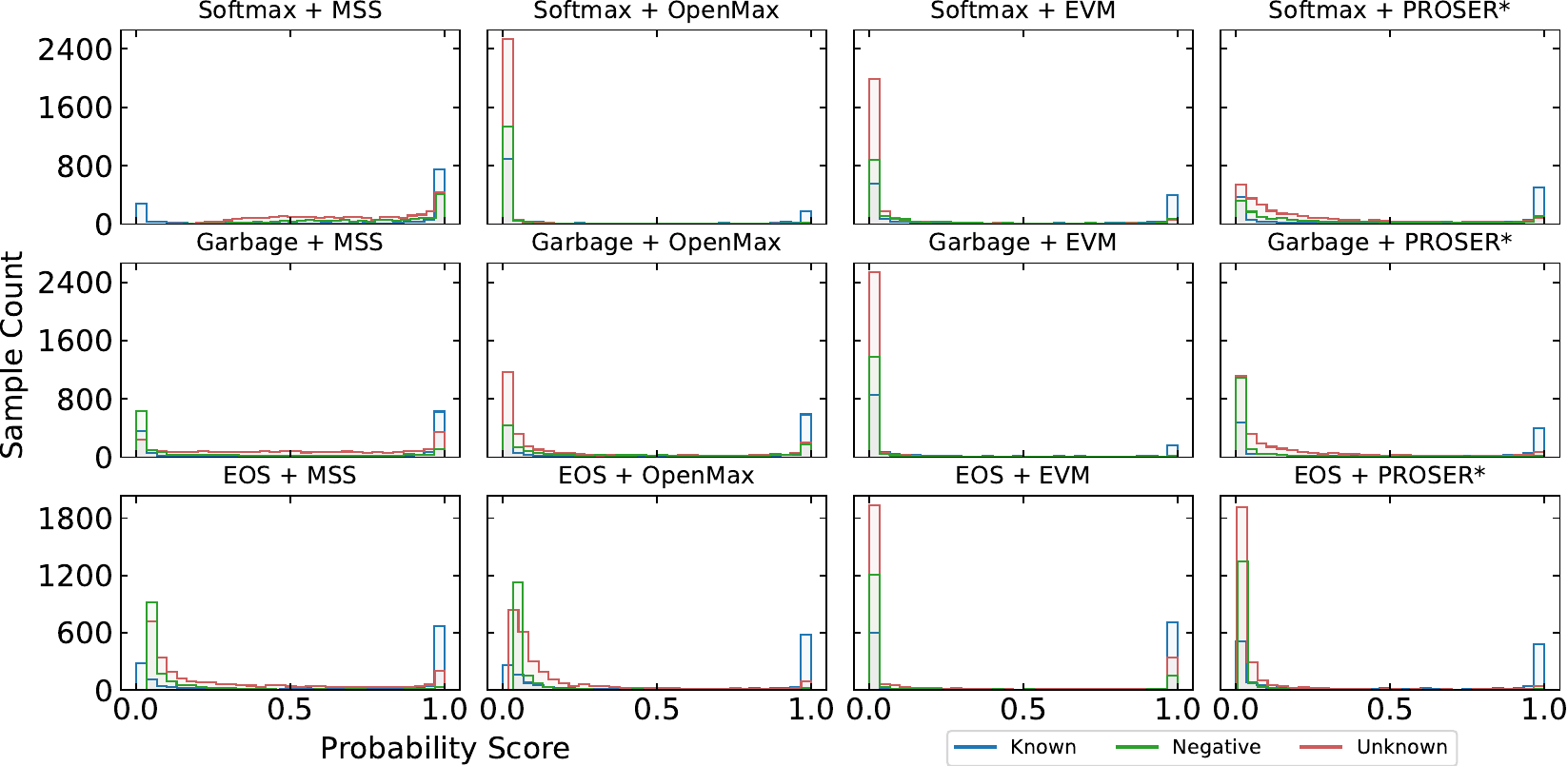}
  }

  \subfloat[Protocol \p3\label{fig:distributions:P3}]{
    \includegraphics[width=.95\textwidth, page=6]{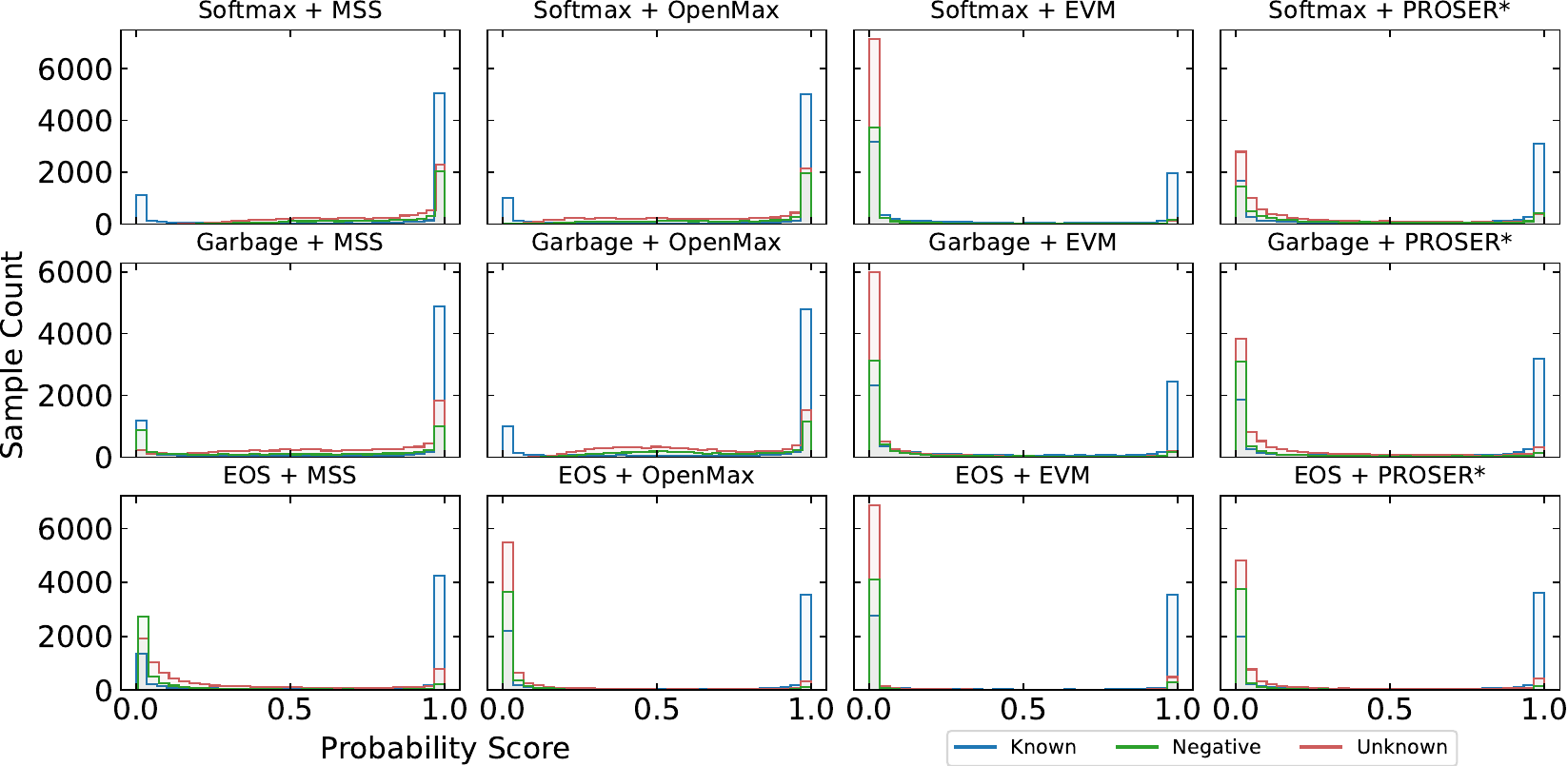}
  }

  \Caption[fig:distributions:others]{Score Distributions}{This figure shows score distributions extracted from the network trained with various loss functions on \subref*{fig:distributions:P2} Protocol \p2 and \subref*{fig:distributions:P3} protocol \p3, and further processed with all algorithms (except for MLS).}
\end{figure}

\newpage
We also include a plot that combines all results. While being busy, this plot improves the comparability across training functions.
\begin{figure}[H]
  \includegraphics[width=.99\textwidth, page=1]{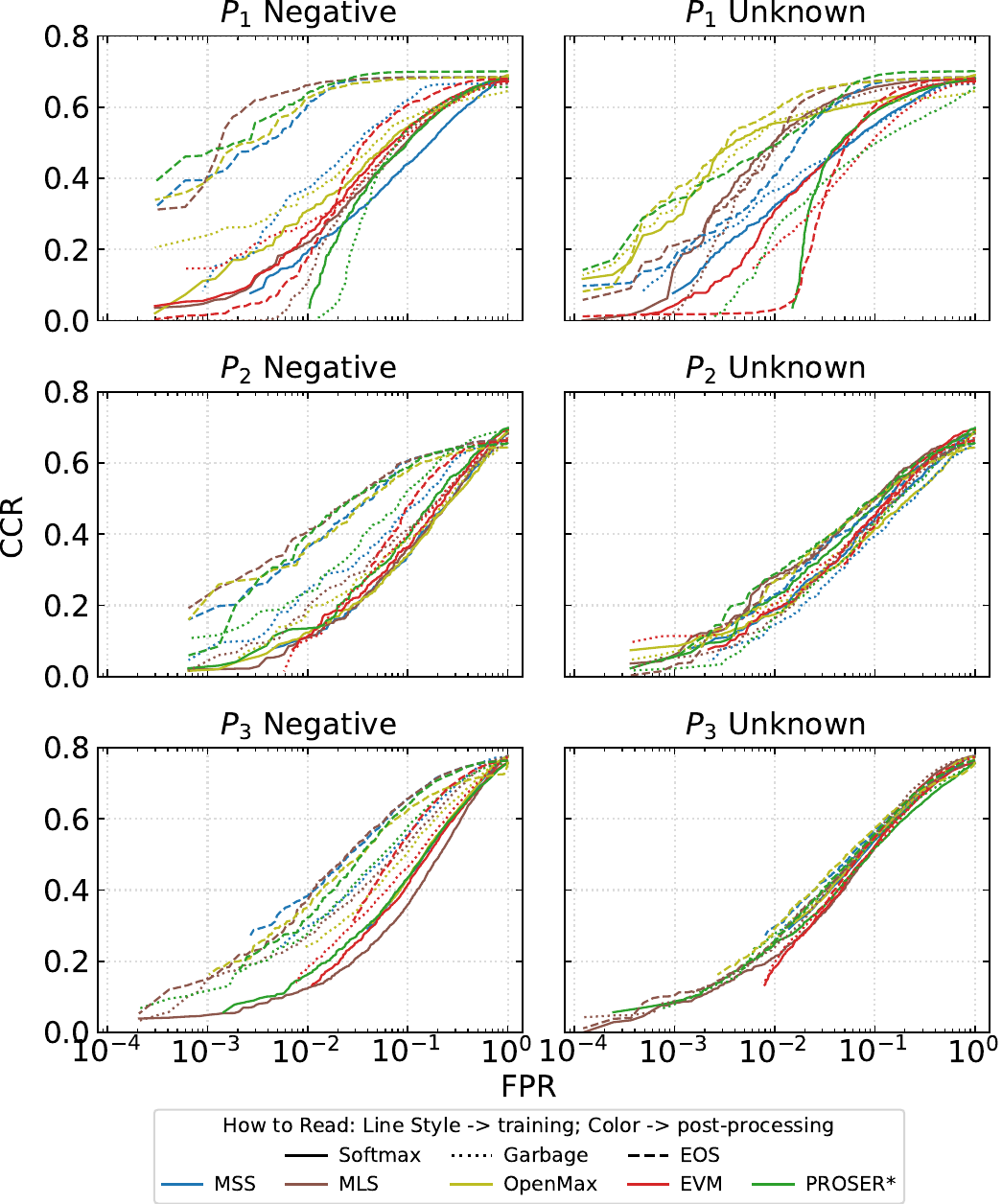}
  \Caption[fig:results]{OSCR Plots}{This figure shows OSCR plots for negative and unknown test samples on all protocols. Line styles distinguish training-based, colors post-processing methods.}
  \label{fig:OSCR-combined}
\end{figure}

\newpage
Additionally, we provide all CCR scores at several FPR values for all parameters of OpenMax, EVM and \proser{}, as obtained on the validation set.
The $\sum$ column includes the sum of all existing CCR values at FPR values of $\zeta=\{10^{-3}, 10^{-2}, 10^{-1}, 1\}$, according to \eqref{eq:ccr@fpr}.
Best values are highlighted in bold font.

\begin{table}[H]
  \tiny\centering
  \subfloat[SoftMax]{
\begin{tabular}{*{3}{|@{\,}c@{\,}}|*{4}{|@{\,}c@{\,}}||@{\,}c@{\,}|}
      \hline
      $\lambda$ & $\kappa$ & $\omega$ & $10^{-3}$ & $10^{-2}$ & $10^{-1}$ & $1$ & $\Sigma$ \\\hline\hline
25 & 0.2 & 1 & 0.0397 & 0.1740 & 0.3977 & 0.5267 & 1.1381 \\\hline
25 & 0.3 & 1 & 0.0606 & 0.1999 & 0.4361 & 0.5707 & 1.2673 \\\hline
25 & 0.4 & 1 & 0.0738 & 0.2227 & 0.4655 & 0.5972 & 1.3592 \\\hline
25 & 0.5 & 1 & \bf 0.0850 & 0.2406 & 0.4918 & 0.6233 & 1.4407 \\\hline
25 & 0.7 & 1 &  & 0.2601 & 0.5076 & 0.6512 & 1.4189 \\\hline
25 & 0.9 & 1 &  &  & 0.4867 & 0.6500 & 1.1367 \\\hline
25 & 1.0 & 1 &  &  &  & 0.6271 & 0.6271 \\
\hline

    \end{tabular}
  }
  \subfloat[Garbage]{
    \begin{tabular}{*{4}{|@{\,}c@{\,}}||@{\,}c@{\,}|}
      \hline
      $10^{-3}$ & $10^{-2}$ & $10^{-1}$ & $1$ & $\Sigma$ \\\hline\hline
0.0046 & 0.0643 & 0.3509 & 0.4867 & 0.9065 \\\hline
0.0115 & 0.0855 & 0.3952 & 0.5308 & 1.0230 \\\hline
0.0300 & 0.1218 & 0.4378 & 0.5650 & 1.1547 \\\hline
0.0663 & 0.1673 & 0.4695 & 0.5951 & 1.2982 \\\hline
 & 0.2580 & 0.4880 & 0.6322 & 1.3782 \\\hline
 &  & 0.4688 & 0.6358 & 1.1045 \\\hline
 &  &  & 0.6140 & 0.6140 \\
\hline

    \end{tabular}
  }
  \subfloat[EOS]{
    \begin{tabular}{*{4}{|@{\,}c@{\,}}||@{\,}c@{\,}|}
      \hline
      $10^{-3}$ & $10^{-2}$ & $10^{-1}$ & $1$ & $\Sigma$ \\\hline\hline
0.0176 & 0.1276 & 0.5134 & 0.5662 & 1.2247 \\\hline
0.0245 & 0.1488 & 0.5371 & 0.6033 & 1.3138 \\\hline
\bf 0.0377 & 0.1750 & 0.4925 & 0.6227 & 1.3279 \\\hline
 & 0.1586 & 0.3657 & 0.6410 & 1.1652 \\\hline
 &  &  & 0.6584 & 0.6584 \\\hline
 &  &  & 0.6601 & 0.6601 \\\hline
 &  &  & 0.6420 & 0.6420 \\
\hline

    \end{tabular}
  }
  \Caption[tab:optimized-evm-1]{EVM Parameter Optimization on \p1}{}
\end{table}

\begin{table}[p]
  \tiny\centering
  \subfloat[SoftMax]{
    \begin{tabular}{*{3}{|@{\,}c@{\,}}|*{4}{|@{\,}c@{\,}}||@{\,}c@{\,}|}
      \hline
      $\lambda$ & $\kappa$ & $\omega$ & $10^{-3}$ & $10^{-2}$ & $10^{-1}$ & $1$ & $\Sigma$ \\\hline\hline
25 & 0.2 & 1 &  & 0.1221 & 0.3542 & 0.6690 & 1.1453 \\\hline
25 & 0.3 & 1 &  & 0.1220 & 0.3596 & 0.6697 & 1.1513 \\\hline
25 & 0.4 & 1 &  &  & 0.3643 & 0.6733 & 1.0377 \\\hline
25 & 0.5 & 1 &  &  & 0.3581 & 0.6739 & 1.0320 \\\hline
25 & 0.7 & 1 &  &  &  & 0.6741 & 0.6741 \\\hline
25 & 0.9 & 1 &  &  &  & 0.6658 & 0.6658 \\\hline
25 & 1.0 & 1 &  &  &  & 0.6531 & 0.6531 \\
\hline

    \end{tabular}
  }
  \subfloat[Garbage]{
    \begin{tabular}{*{4}{|@{\,}c@{\,}}||@{\,}c@{\,}|}
      \hline
      $10^{-3}$ & $10^{-2}$ & $10^{-1}$ & $1$ & $\Sigma$ \\\hline\hline
 & 0.1368 & 0.3771 & 0.6582 & 1.1721 \\\hline
 & 0.1334 & 0.3841 & 0.6631 & 1.1806 \\\hline
 &  & 0.3854 & 0.6675 & 1.0529 \\\hline
 &  & 0.3816 & 0.6719 & 1.0536 \\\hline
 &  &  & 0.6765 & 0.6765 \\\hline
 &  &  & 0.6669 & 0.6669 \\\hline
 &  &  & 0.6501 & 0.6501 \\
\hline

    \end{tabular}
  }
  \subfloat[EOS]{
    \begin{tabular}{*{4}{|@{\,}c@{\,}}||@{\,}c@{\,}|}
      \hline
      $10^{-3}$ & $10^{-2}$ & $10^{-1}$ & $1$ & $\Sigma$ \\\hline\hline
 &  & 0.4484 & 0.6689 & 1.1172 \\\hline
 &  & 0.4399 & 0.6690 & 1.1089 \\\hline
 &  &  & 0.6705 & 0.6705 \\\hline
 &  &  & 0.6710 & 0.6710 \\\hline
 &  &  & \bf 0.6751 & 0.6751 \\\hline
 &  &  & 0.6721 & 0.6721 \\\hline
 &  &  & 0.6678 & 0.6678 \\
\hline

    \end{tabular}
  }
  \Caption[tab:optimized-evm-2]{EVM Parameter Optimization on \p2}{}
\end{table}

\begin{table}[p]
  \tiny\centering
  \subfloat[SoftMax]{
    \begin{tabular}{*{3}{|@{\,}c@{\,}}|*{4}{|@{\,}c@{\,}}||@{\,}c@{\,}|}
      \hline
      $\lambda$ & $\kappa$ & $\omega$ & $10^{-3}$ & $10^{-2}$ & $10^{-1}$ & $1$ & $\Sigma$ \\\hline\hline
25 & 0.2 & 1 &  & 0.1067 & 0.3622 & 0.6296 & 1.0985 \\\hline
25 & 0.3 & 1 &  & 0.1191 & 0.3860 & 0.7038 & 1.2089 \\\hline
25 & 0.4 & 1 &  & 0.1252 & 0.4035 & 0.7208 & 1.2494 \\\hline
25 & 0.5 & 1 &  &  & 0.4171 & 0.7392 & 1.1563 \\\hline
25 & 0.7 & 1 &  &  & \bf 0.4365 & 0.7617 & 1.1983 \\\hline
25 & 0.9 & 1 &  &  &  & 0.7621 & 0.7621 \\\hline
25 & 1.0 & 1 &  &  &  & 0.7368 & 0.7368 \\
\hline

    \end{tabular}
  }
  \subfloat[Garbage]{
    \begin{tabular}{*{4}{|@{\,}c@{\,}}||@{\,}c@{\,}|}
      \hline
      $10^{-3}$ & $10^{-2}$ & $10^{-1}$ & $1$ & $\Sigma$ \\\hline\hline
 & 0.1275 & 0.3629 & 0.5937 & 1.0841 \\\hline
 & 0.1518 & 0.4029 & 0.6607 & 1.2153 \\\hline
 & 0.1711 & 0.4328 & 0.6918 & 1.2957 \\\hline
 & 0.1830 & 0.4540 & 0.7193 & 1.3564 \\\hline
 &  & 0.4751 & 0.7588 & 1.2338 \\\hline
 &  &  & 0.7641 & 0.7641 \\\hline
 &  &  & 0.7357 & 0.7357 \\
\hline

    \end{tabular}
  }
  \subfloat[EOS]{
    \begin{tabular}{*{4}{|@{\,}c@{\,}}||@{\,}c@{\,}|}
      \hline
      $10^{-3}$ & $10^{-2}$ & $10^{-1}$ & $1$ & $\Sigma$ \\\hline\hline
 &  & 0.5474 & 0.7487 & 1.2961 \\\hline
 &  & 0.5509 & 0.7576 & 1.3085 \\\hline
 &  & 0.5497 & 0.7600 & 1.3098 \\\hline
 &  & 0.5482 & 0.7609 & 1.3091 \\\hline
 &  &  & 0.7618 & 0.7618 \\\hline
 &  &  & 0.7602 & 0.7602 \\\hline
 &  &  & 0.7547 & 0.7547 \\
\hline

    \end{tabular}
  }
  \Caption[tab:optimized-evm-3]{EVM Parameter Optimization on \p3}{}
\end{table}

\begin{table}[p]
  \fontsize{4.5}{5.5}\selectfont
  \centering
  \subfloat[SoftMax]{
    \begin{tabular}{*{3}{|@{\,}c@{\,}}|*{4}{|@{\,}c@{\,}}||@{\,}c@{\,}|}
      \hline
      $\lambda$ & $\kappa$ & $\alpha$ & $10^{-3}$ & $10^{-2}$ & $10^{-1}$ & $1$ & $\Sigma$ \\\hline\hline
10 & 1.5 & 2 &  & 0.1908 & 0.4235 & 0.6540 & 1.2683 \\\hline
10 & 1.5 & 3 &  & 0.1911 & 0.4237 & 0.6541 & 1.2689 \\\hline
10 & 1.5 & 5 &  & 0.1911 & 0.4232 & 0.6540 & 1.2683 \\\hline
10 & 1.5 & 10 &  & 0.1911 & 0.4226 & 0.6541 & 1.2678 \\\hline
10 & 1.7 & 2 &  & 0.1908 & 0.4235 & 0.6537 & 1.2679 \\\hline
10 & 1.7 & 3 &  & 0.1911 & 0.4235 & 0.6537 & 1.2683 \\\hline
10 & 1.7 & 5 &  & 0.1911 & 0.4235 & 0.6537 & 1.2683 \\
\hline

    \end{tabular}
  }
  \subfloat[Garbage]{
    \begin{tabular}{*{4}{|@{\,}c@{\,}}||@{\,}c@{\,}|}
      \hline
      $10^{-3}$ & $10^{-2}$ & $10^{-1}$ & $1$ & $\Sigma$ \\\hline\hline
 & 0.1943 & 0.4129 & 0.6397 & 1.2469 \\\hline
 & 0.1943 & 0.4129 & 0.6398 & 1.2469 \\\hline
 & 0.1944 & 0.4124 & 0.6398 & 1.2466 \\\hline
 & 0.1933 & 0.4119 & 0.6399 & 1.2451 \\\hline
 & 0.1944 & 0.4125 & 0.6397 & 1.2466 \\\hline
 & 0.1944 & 0.4125 & 0.6398 & 1.2467 \\\hline
 & 0.1945 & 0.4126 & 0.6398 & 1.2468 \\
\hline

    \end{tabular}
  }
  \subfloat[EOS]{
    \begin{tabular}{*{4}{|@{\,}c@{\,}}||@{\,}c@{\,}|}
      \hline
      $10^{-3}$ & $10^{-2}$ & $10^{-1}$ & $1$ & $\Sigma$ \\\hline\hline
0.2954 & 0.5759 & 0.6601 & 0.6616 & 2.1931 \\\hline
0.2955 & 0.5760 & 0.6602 & 0.6617 & 2.1934 \\\hline
0.2950 & 0.5748 & 0.6601 & 0.6617 & 2.1917 \\\hline
0.2943 & 0.5747 & 0.6602 & 0.6617 & 2.1909 \\\hline
0.2954 & 0.5761 & 0.6602 & 0.6618 & 2.1935 \\\hline
0.2954 & 0.5761 & 0.6602 & 0.6618 & 2.1935 \\\hline
0.2954 & 0.5760 & 0.6602 & 0.6618 & 2.1934 \\
\hline

    \end{tabular}
  }
  \Caption[tab:optimized-openmax-1]{OpenMax Parameter Optimization on \p1}{}

\end{table}

\begin{table}[p]
  \fontsize{4.5}{5.5}\selectfont
  \centering
  \subfloat[SoftMax]{
    \begin{tabular}{*{3}{|@{\,}c@{\,}}|*{4}{|@{\,}c@{\,}}||@{\,}c@{\,}|}
      \hline
      $\lambda$ & $\kappa$ & $\alpha$ & $10^{-3}$ & $10^{-2}$ & $10^{-1}$ & $1$ & $\Sigma$ \\\hline\hline
10 & 1.5 & 2 &  & 0.1065 & 0.3351 & 0.6715 & 1.1131 \\\hline
10 & 1.5 & 3 &  & 0.1083 & 0.3346 & 0.6719 & 1.1148 \\\hline
10 & 1.5 & 5 &  & 0.1071 & 0.3344 & 0.6718 & 1.1134 \\\hline
10 & 1.5 & 10 &  & 0.1070 & 0.3342 & 0.6718 & 1.1130 \\\hline
10 & 1.7 & 2 &  & 0.1065 & 0.3351 & 0.6715 & 1.1131 \\\hline
10 & 1.7 & 3 &  & 0.1070 & 0.3340 & 0.6719 & 1.1130 \\\hline
10 & 1.7 & 5 &  & 0.1081 & 0.3344 & 0.6718 & 1.1143 \\
\hline

    \end{tabular}
  }
  \subfloat[Garbage]{
    \begin{tabular}{*{4}{|@{\,}c@{\,}}||@{\,}c@{\,}|}
      \hline
      $10^{-3}$ & $10^{-2}$ & $10^{-1}$ & $1$ & $\Sigma$ \\\hline\hline
 & 0.1549 & 0.3566 & 0.6582 & 1.1697 \\\hline
 & 0.1549 & 0.3569 & 0.6586 & 1.1704 \\\hline
 & 0.1535 & 0.3566 & 0.6591 & 1.1692 \\\hline
 & 0.1524 & 0.3544 & 0.6591 & 1.1658 \\\hline
 & 0.1549 & 0.3567 & 0.6585 & 1.1701 \\\hline
 & 0.1557 & 0.3558 & 0.6589 & 1.1704 \\\hline
 & 0.1553 & 0.3559 & 0.6585 & 1.1697 \\
\hline

    \end{tabular}
  }
  \subfloat[EOS]{
    \begin{tabular}{*{4}{|@{\,}c@{\,}}||@{\,}c@{\,}|}
      \hline
      $10^{-3}$ & $10^{-2}$ & $10^{-1}$ & $1$ & $\Sigma$ \\\hline\hline
 & 0.3230 & 0.5898 & 0.6645 & 1.5772 \\\hline
 & 0.3278 & 0.5898 & 0.6658 & 1.5835 \\\hline
 & 0.3278 & \bf 0.5901 & 0.6657 & 1.5836 \\\hline
 & 0.3188 & 0.5818 & 0.6663 & 1.5669 \\\hline
 & 0.3221 & 0.5890 & 0.6651 & 1.5763 \\\hline
 & 0.3267 & 0.5897 & 0.6657 & 1.5821 \\\hline
 & 0.3288 & \bf 0.5901 & \bf 0.6672 & 1.5861 \\
\hline

    \end{tabular}
  }
  \Caption[tab:optimized-openmax-2]{OpenMax Parameter Optimization on \p2}{}
\end{table}

\begin{table}[p]
  \fontsize{4.5}{5.5}\selectfont
  \centering
  \subfloat[SoftMax]{
    \begin{tabular}{*{3}{|@{\,}c@{\,}}|*{4}{|@{\,}c@{\,}}||@{\,}c@{\,}|}
      \hline
      $\lambda$ & $\kappa$ & $\alpha$ & $10^{-3}$ & $10^{-2}$ & $10^{-1}$ & $1$ & $\Sigma$ \\\hline\hline
10 & 1.5 & 2 &  &  & 0.4379 & 0.7758 & 1.2137 \\\hline
10 & 1.5 & 3 &  &  & 0.4379 & 0.7759 & 1.2138 \\\hline
10 & 1.5 & 5 &  &  & 0.4371 & 0.7760 & 1.2132 \\\hline
10 & 1.5 & 10 &  &  & 0.4368 & 0.7761 & 1.2129 \\\hline
10 & 1.7 & 2 &  &  & 0.4377 & 0.7757 & 1.2134 \\\hline
10 & 1.7 & 3 &  &  & 0.4378 & 0.7757 & 1.2135 \\\hline
10 & 1.7 & 5 &  &  & 0.4379 & 0.7757 & 1.2136 \\
\hline

    \end{tabular}
  }
  \subfloat[Garbage]{
    \begin{tabular}{*{4}{|@{\,}c@{\,}}||@{\,}c@{\,}|}
      \hline
      $10^{-3}$ & $10^{-2}$ & $10^{-1}$ & $1$ & $\Sigma$ \\\hline\hline
 &  & 0.5078 & 0.7714 & 1.2792 \\\hline
 &  & 0.5077 & 0.7715 & 1.2792 \\\hline
 &  & 0.5077 & 0.7715 & 1.2792 \\\hline
 &  & 0.5080 & 0.7716 & 1.2795 \\\hline
 &  & 0.5078 & 0.7712 & 1.2790 \\\hline
 &  & 0.5078 & 0.7713 & 1.2791 \\\hline
 &  & 0.5079 & 0.7713 & 1.2792 \\
\hline

    \end{tabular}
  }
  \subfloat[EOS]{
    \begin{tabular}{*{4}{|@{\,}c@{\,}}||@{\,}c@{\,}|}
      \hline
      $10^{-3}$ & $10^{-2}$ & $10^{-1}$ & $1$ & $\Sigma$ \\\hline\hline
 & 0.4044 & 0.6621 & 0.7587 & 1.8252 \\\hline
 & 0.4054 & 0.6624 & 0.7585 & 1.8263 \\\hline
 & 0.4042 & 0.6622 & 0.7586 & 1.8250 \\\hline
 & 0.4044 & 0.6639 & 0.7587 & 1.8270 \\\hline
 & 0.4045 & 0.6618 & 0.7596 & 1.8258 \\\hline
 & 0.4049 & 0.6618 & 0.7593 & 1.8260 \\\hline
 & 0.4061 & 0.6618 & 0.7591 & 1.8270 \\
\hline

    \end{tabular}
  }
  \Caption[tab:optimized-openmax-3]{OpenMax Parameter Optimization on \p3}{}
\end{table}

\begin{table}[t!]
  \fontsize{4.5}{5.5}\selectfont
  \centering
  \subfloat[SoftMax]{
    \begin{tabular}{|@{\,}c@{\,}|*{4}{|@{\,}c@{\,}}||@{\,}c@{\,}|}
      \hline
      $B$ & $10^{-3}$ & $10^{-2}$ & $10^{-1}$ & $1$ & $\Sigma$ \\\hline\hline
1 &  &  & 0.3854 & 0.6075 & 0.9928 \\\hline
2 &  &  & 0.4628 & 0.6486 & 1.1114 \\\hline
5 &  & \bf 0.0336 & 0.3941 & 0.6136 & 1.0413 \\\hline
10 &  &  & 0.4570 & 0.6392 & 1.0962 \\\hline
25 &  & 0.0292 & 0.4067 & 0.6200 & 1.0559 \\\hline
100 &  &  & \bf 0.4809 & \bf 0.6532 & \bf 1.1342 \\\hline

    \end{tabular}
  }
  \subfloat[Garbage]{
    \begin{tabular}{*{4}{|@{\,}c@{\,}}||@{\,}c@{\,}|}
      \hline
      $10^{-3}$ & $10^{-2}$ & $10^{-1}$ & $1$ & $\Sigma$ \\\hline\hline
 &  & 0.3894 & 0.6331 & 1.0225 \\\hline
 &  & \bf 0.5123 & 0.6321 & \bf 1.1445 \\\hline
 &  & 0.4352 & 0.6180 & 1.0532 \\\hline
 &  & 0.4987 & 0.6243 & 1.1230 \\\hline
 &  & 0.4803 & \bf 0.6386 & 1.1189 \\\hline
 &  & 0.4976 & 0.6288 & 1.1264 \\\hline

    \end{tabular}
  }
  \subfloat[EOS]{
    \begin{tabular}{*{4}{|@{\,}c@{\,}}||@{\,}c@{\,}|}
      \hline
      $10^{-3}$ & $10^{-2}$ & $10^{-1}$ & $1$ & $\Sigma$ \\\hline\hline
0.2334 & 0.3859 & 0.6671 & 0.6744 & 1.9608 \\\hline
0.2937 & 0.4894 & 0.6610 & 0.6728 & 2.1168 \\\hline
\bf 0.4011 & \bf 0.5860 & \bf 0.6744 & \bf 0.6789 & \bf 2.3405 \\\hline
0.0428 & 0.3809 & 0.6074 & 0.6499 & 1.6810 \\\hline
0.0666 & 0.2911 & 0.6165 & 0.6551 & 1.6293 \\\hline
0.1803 & 0.3756 & 0.5711 & 0.5755 & 1.7025 \\\hline

    \end{tabular}
  }
  \Caption[tab:optimized-proser-1]{\proser{} Parameter Optimization on \p1}{}

\end{table}

\begin{table}[t!]
  \fontsize{4.5}{5.5}\selectfont
  \centering
  \subfloat[SoftMax]{
    \begin{tabular}{|@{\,}c@{\,}|*{4}{|@{\,}c@{\,}}||@{\,}c@{\,}|}
      \hline
      $B$ & $10^{-3}$ & $10^{-2}$ & $10^{-1}$ & $1$ & $\Sigma$ \\\hline\hline
1 & \bf 0.0284 & 0.1282 & \bf 0.3846 & 0.6899 & \bf 1.2310 \\\hline
2 & \bf 0.0284 & \bf 0.1355 & 0.3713 & 0.6912 & 1.2263 \\\hline
5 & 0.0233 & 0.1071 & 0.3587 & 0.6923 & 1.1813 \\\hline
10 & 0.0206 & 0.1026 & 0.3364 & 0.6895 & 1.1491 \\\hline
25 & 0.0203 & 0.0962 & 0.3393 & \bf 0.6945 & 1.1503 \\\hline
100 & 0.0145 & 0.0825 & 0.3415 & 0.6910 & 1.1296 \\\hline

    \end{tabular}
  }
  \subfloat[Garbage]{
    \begin{tabular}{*{4}{|@{\,}c@{\,}}||@{\,}c@{\,}|}
      \hline
      $10^{-3}$ & $10^{-2}$ & $10^{-1}$ & $1$ & $\Sigma$ \\\hline\hline
0.0537 & 0.2339 & \bf 0.5101 & 0.6669 & 1.4647 \\\hline
0.0590 & 0.2349 & 0.4993 & 0.6700 & 1.4632 \\\hline
0.0570 & \bf 0.2483 & 0.5033 & \bf 0.6775 & \bf 1.4862 \\\hline
\bf 0.0745 & 0.2294 & 0.4976 & 0.6773 & 1.4788 \\\hline
0.0504 & 0.2283 & 0.4938 & 0.6712 & 1.4437 \\\hline
0.0594 & 0.2208 & 0.5017 & 0.6766 & 1.4585 \\\hline

    \end{tabular}
  }
  \subfloat[EOS]{
    \begin{tabular}{*{4}{|@{\,}c@{\,}}||@{\,}c@{\,}|}
      \hline
      $10^{-3}$ & $10^{-2}$ & $10^{-1}$ & $1$ & $\Sigma$ \\\hline\hline
0.0662 & 0.3113 & 0.5759 & 0.6491 & 1.6024 \\\hline
0.0538 & 0.2569 & 0.5534 & 0.6406 & 1.5048 \\\hline
 & 0.3169 & 0.5739 & 0.6495 & 1.5403 \\\hline
\bf 0.0911 & \bf 0.3270 & 0.5717 & 0.6523 & \bf 1.6420 \\\hline
 & 0.3210 & \bf 0.5880 & \bf 0.6571 & 1.5662 \\\hline
0.0545 & 0.3000 & 0.5850 & 0.6552 & 1.5947 \\\hline

    \end{tabular}
  }
  \Caption[tab:optimized-proser-2]{\proser{} Parameter Optimization on \p2}{}
\end{table}

\begin{table}[t!]
  \fontsize{4.5}{5.5}\selectfont
  \centering
  \subfloat[SoftMax]{
    \begin{tabular}{|@{\,}c@{\,}|*{4}{|@{\,}c@{\,}}||@{\,}c@{\,}|}
      \hline
      $B$ & $10^{-3}$ & $10^{-2}$ & $10^{-1}$ & $1$ & $\Sigma$ \\\hline\hline
1 &  & 0.1472 & 0.4228 & 0.7466 & 1.3166 \\\hline
2 &  & 0.1446 & 0.4342 & 0.7500 & 1.3288 \\\hline
5 &  & 0.1430 & 0.4267 & 0.7451 & 1.3148 \\\hline
10 &  & \bf 0.1545 & \bf 0.4387 & \bf 0.7524 & \bf 1.3456 \\\hline
25 &  & 0.1506 & 0.4259 & 0.7465 & 1.3230 \\\hline
100 &  & 0.1494 & 0.4271 & 0.7453 & 1.3218 \\\hline

    \end{tabular}
  }
  \subfloat[Garbage]{
    \begin{tabular}{*{4}{|@{\,}c@{\,}}||@{\,}c@{\,}|}
      \hline
      $10^{-3}$ & $10^{-2}$ & $10^{-1}$ & $1$ & $\Sigma$ \\\hline\hline
 & 0.2725 & 0.5505 & 0.7455 & 1.5686 \\\hline
 & 0.2881 & 0.5722 & \bf 0.7583 & 1.6186 \\\hline
 & 0.2813 & 0.5610 & 0.7471 & 1.5894 \\\hline
0.1143 & \bf 0.2967 & \bf 0.5746 & 0.7539 & \bf 1.7394 \\\hline
0.1079 & 0.2918 & 0.5654 & 0.7464 & 1.7115 \\\hline
\bf 0.1175 & 0.2864 & 0.5629 & 0.7478 & 1.7145 \\\hline

    \end{tabular}
  }
  \subfloat[EOS]{
    \begin{tabular}{*{4}{|@{\,}c@{\,}}||@{\,}c@{\,}|}
      \hline
      $10^{-3}$ & $10^{-2}$ & $10^{-1}$ & $1$ & $\Sigma$ \\\hline\hline
 &  & 0.4485 & 0.6855 & 1.1340 \\\hline
 & 0.2211 & 0.6056 & 0.7380 & 1.5647 \\\hline
 & 0.2963 & \bf 0.6431 & 0.7584 & 1.6977 \\\hline
 & 0.1730 & 0.6240 & 0.7499 & 1.5469 \\\hline
 &  & 0.6110 & 0.7430 & 1.3539 \\\hline
 & \bf 0.3251 & 0.6390 & \bf 0.7585 & \bf 1.7226 \\\hline

    \end{tabular}
  }
  \Caption[tab:optimized-proser-3]{\proser{} Parameter Optimization on \p3}{}
\end{table}

\end{document}